\documentclass{article}

\usepackage{microtype}
\usepackage{graphicx}
\usepackage{subcaption}
\usepackage{booktabs}

\usepackage{hyperref}

\usepackage[preprint]{icml2026}

\usepackage{amsmath}
\usepackage{amssymb}
\usepackage{mathtools}
\usepackage{amsthm}

\usepackage[capitalize,noabbrev]{cleveref}

\usepackage{booktabs}
\usepackage{multirow}
\usepackage{amsfonts}
\usepackage{amssymb}

\usepackage{nicefrac}
\usepackage{url}
\usepackage[switch]{lineno}
\usepackage{tikz}
\usepackage{wrapfig}

\makeatletter
\newcommand{\ssymbol}[1]{$^{\@fnsymbol{#1}}$}
\makeatother

\usepackage[misc]{ifsym}
 
\theoremstyle{plain}

\theoremstyle{definition}

\theoremstyle{remark}

\usepackage[textsize=tiny]{todonotes}

\icmltitlerunning{Submission and Formatting Instructions for ICML 2026}

\begin{document}

\twocolumn[
  \icmltitle{RiskAgent: Synergizing Language Models with Validated Tools for Evidence-Based Risk Prediction}

  \icmlsetsymbol{equal}{*}

  \begin{icmlauthorlist}
    \icmlauthor{Fenglin Liu}{equal,ox}
    \icmlauthor{Jinge Wu}{equal,ucl}
    \icmlauthor{Hongjian Zhou}{ox}
    \icmlauthor{Xiao Gu}{ox}
    \icmlauthor{Jiayuan Zhu}{ox}
    \icmlauthor{Jiazhen Pan}{tum}
    \icmlauthor{Junde Wu}{ox}
    \icmlauthor{Soheila Molaei}{ox}
    \icmlauthor{Anshul Thakur}{ox}
    \icmlauthor{Lei Clifton}{ox}
    \icmlauthor{Honghan Wu}{ucl,glasgow}
    \icmlauthor{David Clifton}{ox,oscar}
  \end{icmlauthorlist}
  \icmlaffiliation{ox}{University of Oxford}
  \icmlaffiliation{ucl}{University College London}
  \icmlaffiliation{tum}{Technical University of Munich}
  \icmlaffiliation{glasgow}{University of Glasgow}
  \icmlaffiliation{oscar}{Oxford-Suzhou Centre for Advanced Research}

  \icmlcorrespondingauthor{Fenglin  Liu}{fenglin.liu@eng.ox.ac.uk}

  \icmlkeywords{Machine Learning, ICML}

  \vskip 0.3in
]

\printAffiliationsAndNotice{} 

\begin{abstract}

Large Language Models (LLMs) achieve competitive results compared to human experts in medical examinations. However, it remains a challenge to apply LLMs to complex clinical decision-making, which requires a deep understanding of medical knowledge and differs from the standardized, exam-style scenarios commonly used in current efforts. A common approach is to fine-tune LLMs for target tasks, which, however, not only requires substantial data and computational resources but also remains prone to generating `hallucinations'. In this work, we present RiskAgent, which synergizes language models with hundreds of validated clinical decision tools supported by evidence-based medicine, to provide generalizable and faithful recommendations. Our experiments show that RiskAgent not only achieves superior performance on a broad range of clinical risk predictions across diverse scenarios and diseases, but also demonstrates robust generalization in tool learning on the external MedCalc-Bench dataset, as well as in medical reasoning and question answering on three representative benchmarks—MedQA, MedMCQA, and MMLU.

\end{abstract}

\section{Introduction}
Inspired by the success of Large Language Models (LLMs) \cite{hurst2024gpt4o,jaech2024openaio1,chowdhery2022palm}, an increasing amount of research is attempting to apply LLMs to the medical field and explore their potential in assisting healthcare professionals, resulting in different types of medical LLMs \cite{zhou2023survey}. For example, based on the LLMs PaLM \cite{chowdhery2022palm} and Gemini \cite{google2023bard}, Google has developed MedPaLM-2 \cite{medpalm2} and Med-Gemini \cite{yang2024advancing}, respectively, which achieve 86.5\% and 91.1\% accuracy comparable to the 87.0\% accuracy of human experts \cite{wu2024pmc} on the US Medical Licensing Examination. Based on open-source LLM LLaMA \cite{llama,touvron2023llama2}, dozens of medical LLMs have been proposed, such as Clinical Camel \cite{toma2023clinical} and Meditron \cite{chen2023meditron}, to address different medical tasks.

\begin{figure*}[!htbp]
\centering
\includegraphics[width=1\linewidth]{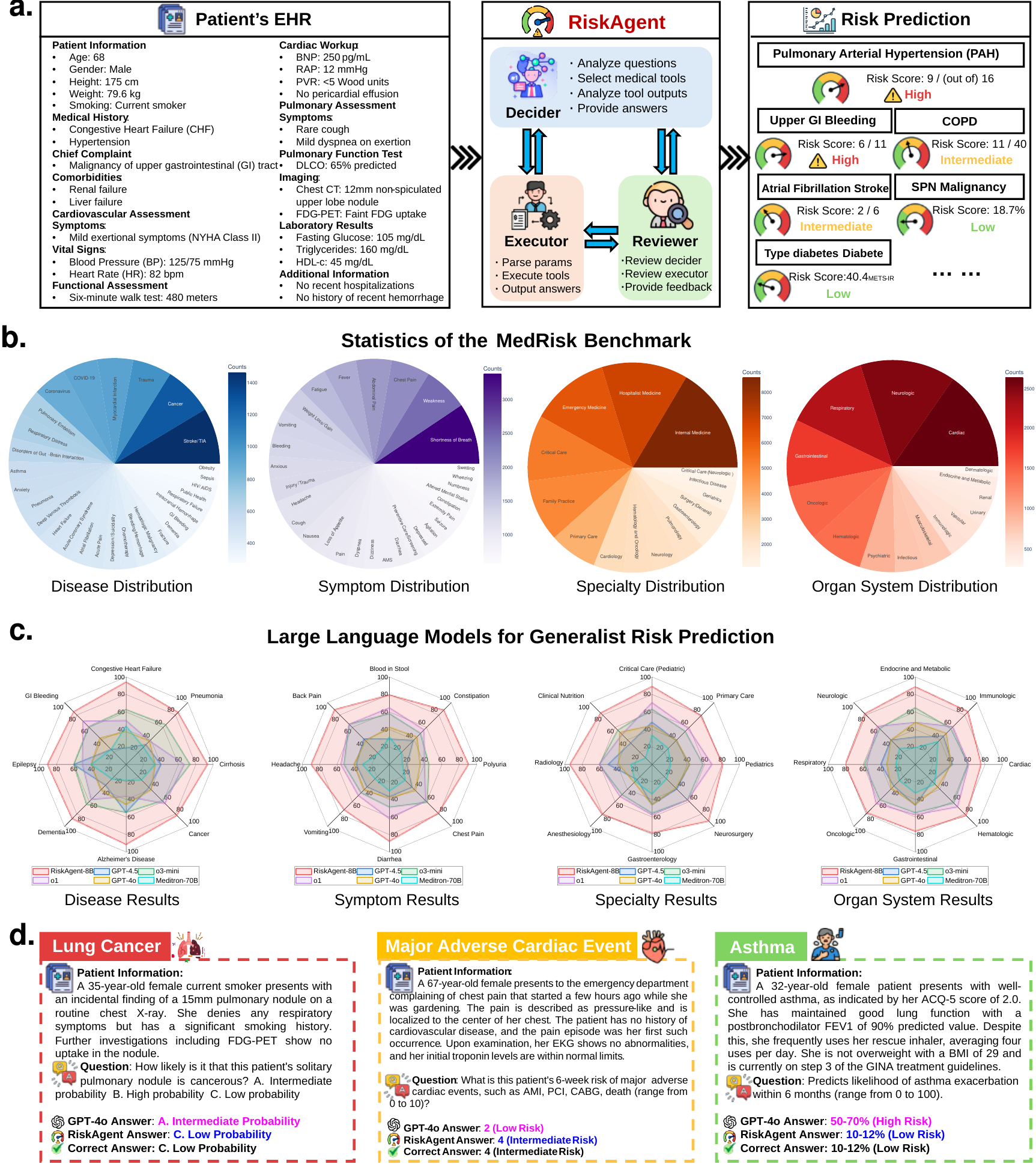}
\caption{\textbf{a.} The RiskAgent, including three LLM agents (Decider, Executor, and Reviewer) can perform multiple medical risk predictions given the patient's healthcare information. 
\textbf{b.} Statistics of the MedRisk benchmark, which consists of 12,352 medical risk questions, covering 154 diseases, 86 symptoms, 50 specialties, and 24 organ systems. 
\textbf{c.} With only 8 billion parameters, our RiskAgent outperforms both existing high-performance general LLMs (i.e., GPT-4o \cite{hurst2024gpt4o}, o1 \cite{jaech2024openaio1}, o3-mini \cite{gpt2025o3}) and state-of-the-art medical LLM (i.e., Meditron-70B) by large margins across different diseases, symptoms, specialties, and organ systems.
In contrast to existing LLMs, RiskAgent can collaborate with evidence-based medical tools to not only substantially increase its risk prediction accuracy, but also deliver evidence-based answers.
The t-tests between the results from RiskAgent and the best-performing LLMs indicate that the improvement is significant with $p < 0.01$.
\textbf{d.} The examples of risk predictions by our method for cancer, cardiac events, and asthma, demonstrate greater accuracy than GPT-4o. The pink- and blue-colored text indicates the incorrect and desirable answers, respectively.
The correct answers are verified by clinicians.
}
\label{fig:introduction}
\end{figure*}
However, the effective application of LLMs to complex medical tasks remains challenging: 1) \textit{Clinical inefficacy}: LLMs achieve superior performance mainly in examination tasks \cite{liu2024large}. However, existing LLMs, including GPT-4, perform poorly on tasks closer to clinical decision-making, such as medical code querying \cite{soroush2024large};
2) \textit{Resource-intensive}: To enable LLMs to deliver accurate results on different tasks, a solution is fine-tuning LLMs on target medical data \cite{van2024adapted,zhou2023survey}. However, fine-tuning LLMs (forcing them to learn all tasks) not only places significant learning pressure on them, but also inevitably requires substantial data (ranging from 1 million \cite{ferber2024gpt,han2023medalpaca,medpalm2} to 80 billion tokens \cite{chen2023meditron,wu2024pmc}) and computational resources that may be prohibitively difficult to obtain for resource-limited medical institutions;
3) \textit{Privacy concern}: An alternative solution is to prompt commercial LLMs (e.g., the GPT series \cite{nori2023medprompt}). However, the use of commercial LLMs involves strict privacy regulations regarding sensitive patient information in the real world \cite{ullah2024challenges}.
4) \textit{Unfaithful outputs}: LLMs still face the widely known `hallucination' challenge \cite{pal2023med,Hallu-survey}: the produced answers appear reasonable but are not based on factual information and knowledge. Furthermore, existing LLMs could not effectively and accurately provide evidence to show the sources of generated answers, which is crucial in healthcare \cite{kitamura2023chatgpt,zhou2023survey}.
Therefore, developing different medical LLMs for diverse medical tasks is expensive, time-consuming, and energy-intensive.

In this work, we present RiskAgent for solving complex medical problems and take risk prediction as a representative example that requires LLMs to not only accurately understand complex patients' health records, but also predict potential health risks, including the risk of developing diseases or the mortality/survival rate of diseases.
This is critical for preventative health \cite{pudjihartono2022review}, since early and accurate risk prediction can alert physicians and patients for early intervention, and thus improve clinical outcomes, especially for complex diseases such as cardiovascular disease and rare diseases \cite{damen2016prediction}.
Figure~\ref{fig:introduction}(a) shows that our RiskAgent includes three LLM agents: Decider, Executor, and Reviewer: 1) The Decider analyzes medical problems and accurately selects appropriate tools from hundreds of options; 2) The Executor understands the selected tools, parses their required parameters, and executes them; 3) The Decider then analyzes the returned tool outputs and provides initial answers; 4) The Executor structures and executes outputs to generate final answers; 5) The Reviewer finally reviews the decision-making process and provides reflection on the results.

To comprehensively evaluate LLMs' risk prediction performance for diverse scenarios, we first build a risk prediction benchmark MedRisk, including 154 diseases, 86 symptoms, 50 specialties, and 24 organ systems, totaling 12,352 cases. Figure \ref{fig:introduction}(b) presents the benchmark's statistics.
We evaluate 19 state-of-the-art methods, covering both general and medical LLMs, as well as open-source public and commercial LLMs.
Figure \ref{fig:introduction}(c) and Table \ref{tab:results} reveal that current LLMs perform poorly in making accurate risk predictions (15.83\%$\sim$58.77\% accuracy) for complex diseases, while our RiskAgent, with 8 billion model parameters, outperforms existing LLMs with large margins.
Figure \ref{fig:introduction}(d) provides three examples showing that our method makes more accurate risk predictions than GPT-4o \cite{hurst2024gpt4o} across different complex diseases.

Overall, the main contributions of our work are:
\begin{itemize}
    \item We present RiskAgent, a clinically efficient multi-agent framework that synergizes language models with diverse validated tools to provide faithful, generalizable, and evidence-based recommendations.

    \item We construct MedRisk, a comprehensive benchmark for generalist medical risk prediction. It encompasses 12,352 distinct clinical questions spanning 154 diseases, 86 symptoms, and 50 specialties.
        
    \item We conduct extensive experiments showing that RiskAgent outperforms state-of-the-art models and exhibits generalization capabilities in risk prediction, tool learning, medical reasoning, and question answering.
\end{itemize}

\section{Related Work}

\paragraph{Medical Large Language Models} The adaptation of general LLMs to the medical domain has gained significant attention~\cite{zhou2023survey}. While these models demonstrate proficiency in standardized examinations, they frequently struggle with real-world clinical tasks requiring high fidelity and precision~\cite{soroush2024large}. A significant limitation is their reliance on internal parametric knowledge for tasks that demand precise calculations or adherence to updated clinical guidelines. This approach ignores the utility of existing deployed medical tools—such as devices, models, and APIs—leading not only to resource waste but also to potential hallucinations and a disconnect from evidence-based practice~\cite{sackett1997evidence}. Instead of attempting to embed all medical knowledge into model weights, our work focuses on enabling LLMs to collaborate with validated evidence-based medical tools.

\vspace{-10pt}
\paragraph{Tool-Augmented Medical Agents} Recent works have explored augmenting medical LLMs with external tools to enhance clinical capabilities. 1) \textit{Multi-agent reasoning}: MedAgents~\cite{tang2024medagents} and MDAgents~\cite{kim2024mdagents} simulate clinical discussions by coordinating multiple agents with different medical roles to improve diagnostic reasoning. While effective for complex decision-making, these systems still rely on parametric knowledge without leveraging validated clinical calculators. 2) \textit{General tool calling}: Some approaches equip medical LLMs with general-purpose tools such as code interpreters or search engines to assist clinical tasks. However, these generic tools lack the domain-specific validation required for precise medical risk assessment. 3) \textit{Clinical calculator integration}: AgentMD~\cite{jin2024agentmd} pioneers connecting LLMs with clinical calculators, demonstrating the value of evidence-based tools. However, it focuses on limited tool coverage and relies on extensive prompting of proprietary models. Our RiskAgent advances this direction by training a lightweight model that collaborates with hundreds of validated tools, reducing resource costs while producing faithful outputs that clinicians can efficiently verify.

\begin{figure*}[t]
\centering
\includegraphics[width=0.85\linewidth]{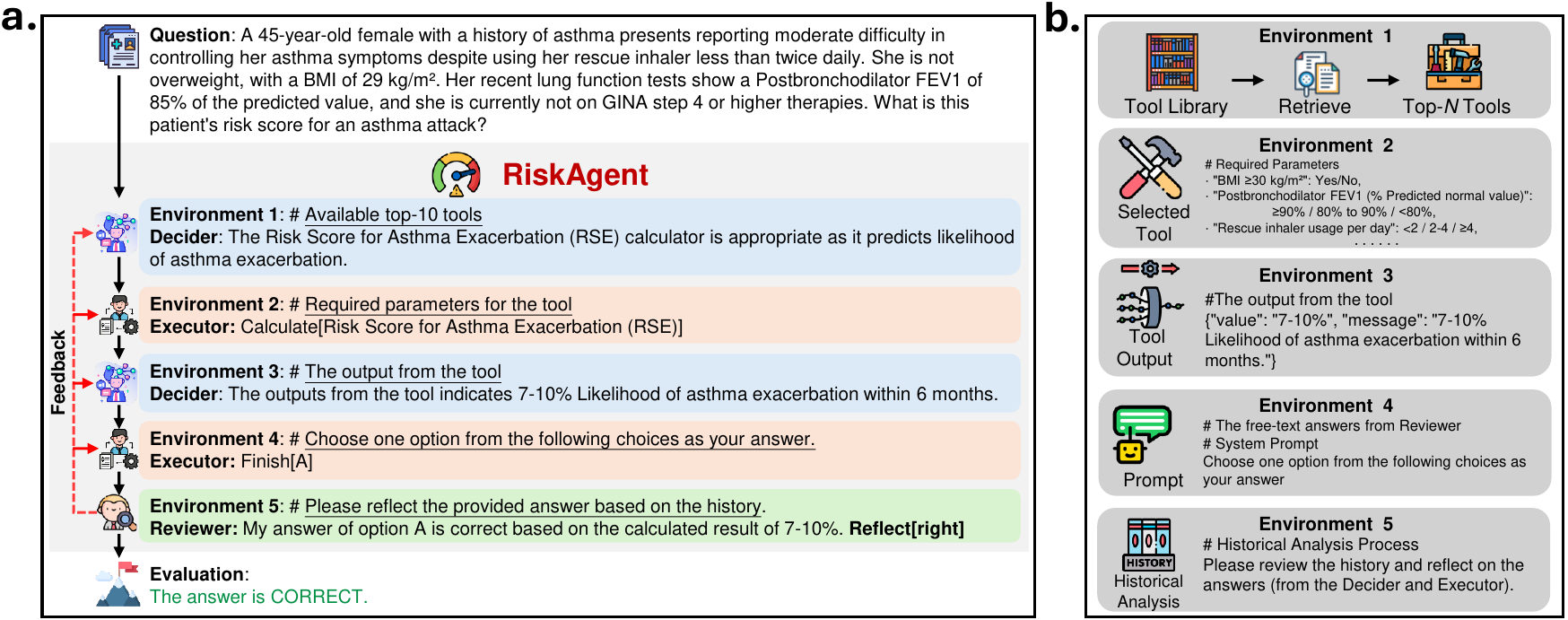}
\caption{Flowchart of the RiskAgent system. \textbf{a.} Data flow in the system. \textbf{b.} Demonstration of the Environment component in the system.
}
\label{fig:framework}
\vspace{-5pt}
\end{figure*}

\section{Approach}
Figure~\ref{fig:framework} shows the flowchart of our RiskAgent system\footnote{
We attach a detailed description of our method in Appendix~\ref{appen:method}.}.
Inspired by the success of actor-critic reinforcement learning \cite{grondman2012survey} that introduces the Critic, Actor, Reward, and Environment, we introduce three LLM agents and one component, i.e., Decider, Executor, Reviewer, and Environment. These modules work together to ensure that our system runs smoothly and accurately.

\subsection{LLM Agents}
\textbf{Decider}: The LLM is instructed to analyze medical problems, decide which medical tools to invoke, analyze the output of the tools, and provide answers.
    
\textbf{Executor}: The LLM is instructed to execute the Decider's decisions, i.e., parse the required parameters to invoke the tools, and convert the answers into risk scores or choices.

\textbf{Reviewer}: The LLM is instructed to review all historical information and the provided answers, and then offer reflection results.

\paragraph{Decider-Executor}
The Decider and Executor in the system are responsible for assigning and executing instructions, respectively. Their collaborative process is as follows:
1) The Decider, as shown in Figure~\ref{fig:framework}, analyzes the given medical problem and the retrieved tools to determine which tools to use;
2) The Executor executes the Decider’s decisions (i.e., the determined tools). In detail, the Executor analyzes the medical problem according to the tools' parameter requirements, parses the necessary parameters accurately, sends them to the tools for invocation, and obtains the tools' results;
3) The Decider further analyzes the tools’ results reported by the Executor and generates answers to the input medical problems;
4) The Executor finally converts the generated answers into risk scores or risk groups as the system outputs.
As we can see, our system reduces the learning burden of each component by dividing responsibilities between the Decider and Executor, thus enabling accurate and efficient collaboration with medical tools.

\paragraph{Reviewer-Decider/Executor}
We further introduce the Reviewer that evaluates the provided answer by considering the historical analysis process and providing reflection results to improve the system's performance.
If the Reviewer determines that the answer is correct, i.e., Reflect[right], the system outputs the answer; otherwise, i.e., Reflect[wrong], the Reviewers provides reflection results to the Reviewer and Executor, and the system continues running. In the case of the Decider selecting the wrong tools, the Reviewer provides reflection results on the Decider's decisions, and the Decider re-selects the tools. The Reviewer also handles the situation when the Executor fails to correctly implement the Decider's decisions, such as a parameter parsing failure. The Reviewer also reviews the progress of the Executor's work to ensure the tool invocation is completed successfully.

\subsection{Environment}
It provides the system with external information that can be interacted with, such as the medical tool library, required parameters for the tools, tool outputs, and system prompts.
We introduce five environments to enhance synergy between LLMs and tools.
 \textit{Environment 1:}
    Hundreds of medical tools are often involved in real-world clinical scenarios, due to their complexity and diversity. To enable LLMs to collaborate with these tools, we introduce a retrieval-ranking algorithm to extract the $N$ most relevant medical tools from a tool library containing $M$ tools based on the current medical problem.
 \textit{Environment 2:}
    When the Decider has chosen a tool, this environment provides the Executor with the required parameters for tool invocation, enabling the Executor to use the tools smoothly and accurately.
  \textit{Environment 3:}
    The returned results from the medical tools are stored in this environment, where the Decider interacts with for analysis and the generation of free-text answers to questions.
  \textit{Environment 4:}
    This environment provides the system prompts to assist the Executor in formatting and converting the free-text answers generated by the Decider into risk scores or choices for evaluation.
  \textit{Environment 5:}
   This environment stores the historical analysis process from the Decider and Executor, as well as system prompts that instruct the Reviewer to perform the review process accurately.

\begin{table*}[t]
\centering
\scriptsize
\caption{Performance of risk prediction on the MedRisk benchmark. `B': Billion.  We use accuracy to report our results.
We conduct five runs for our RiskAgent, GPT-4o, and o3-mini. We report the mean and standard deviation$_\text{(STD)}$ of performance.
In this paper, all values are reported in percentage (\%). Higher is better for all columns. The bold number denotes the best result across all methods.
We select the most common cases across disease, symptom, specialty, and organ system for a broad evaluation and subsequently report the overall results (i.e., the average performance on all test data).  
}
\label{tab:results}
\setlength{\tabcolsep}{0.1pt}

\begin{tabular}{clllllllllll}
\toprule
\multirow{3}{*}[-3pt]{\textbf{Types}} &  \multirow{3}{*}[-3pt]{\textbf{Methods}}   &   \multicolumn{5}{c}{\textbf{MedRisk-Qualitative}}    &   \multicolumn{5}{c}{\textbf{MedRisk-Quantitative}}

\\ \cmidrule(r){3-7} \cmidrule(r){8-12}

&   
& \begin{tabular}[c]{@{}c@{}}Stroke/ \\ TIA \end{tabular}\includegraphics[width=0.4cm]{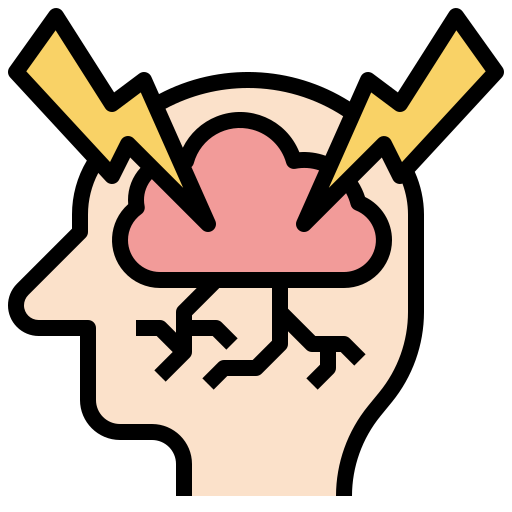} 
& \begin{tabular}[c]{@{}c@{}}Shortness \\of Breath
 \end{tabular}\includegraphics[width=0.4cm]{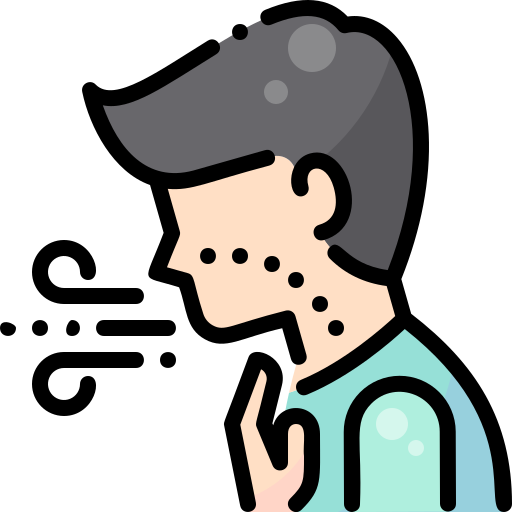} 
&  \begin{tabular}[c]{@{}c@{}}Internal \\Medicine
\end{tabular}\includegraphics[width=0.4cm]{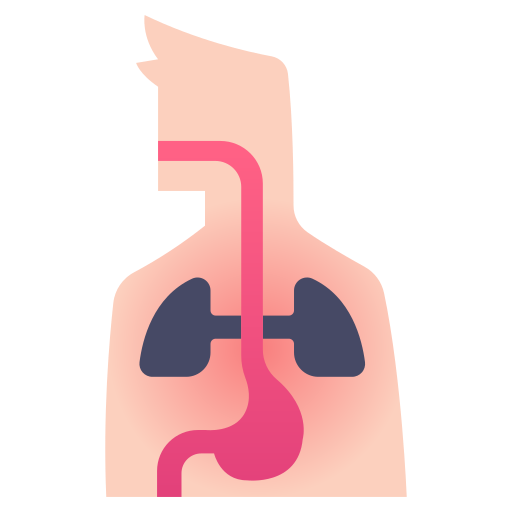}
& \begin{tabular}[c]{@{}c@{}}Cardiac
 \end{tabular}\includegraphics[width=0.4cm]{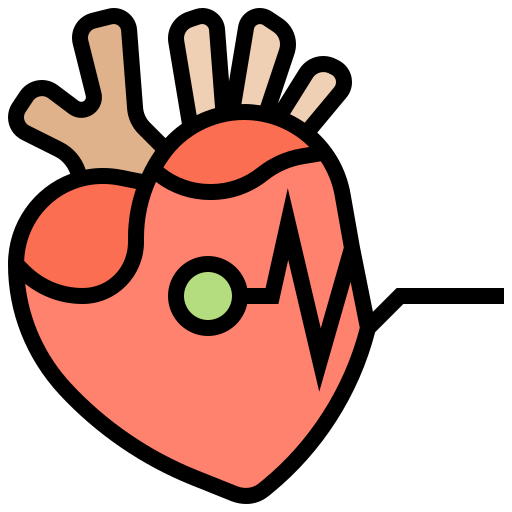}
& \textbf{Overall} 

& \begin{tabular}[c]{@{}c@{}}Stroke/ \\ TIA \end{tabular}\includegraphics[width=0.4cm]{icons/stroke.png}
& \begin{tabular}[c]{@{}c@{}}Shortness \\of Breath
 \end{tabular}\includegraphics[width=0.4cm]{icons/breathing.png}
&  \begin{tabular}[c]{@{}c@{}}Internal \\Medicine
\end{tabular}\includegraphics[width=0.4cm]{icons/internal.png}
& \begin{tabular}[c]{@{}c@{}}Cardiac
 \end{tabular}\includegraphics[width=0.4cm]{icons/heart.png}
& \textbf{Overall} 
\\
\midrule 

\multirow{6}{*}{\rotatebox{90}{\begin{tabular}[c]{@{}c@{}} Public \\ LLMs \end{tabular}}} 

& Mistral-7B \cite{jiang2023mistral}   & 
47.89&
29.75&
33.72&
36.29&
35.47&  
21.13&
23.73&
25.55&
22.39& 
24.35
\\

& LLaMA-3-8B \cite{llama3modelcard} & 
48.59&
44.62&
42.59&
46.33&  
43.75& 
28.87&
27.85&
32.79&
32.05& 
32.06

\\

& Mixtral-8x7B \cite{jiang2024mixtral} & 
44.37&
29.43&
34.19&
37.07& 
34.42& 
16.90&
17.09&
19.84&
17.37&
18.83\\

&  LLaMA-3-70B \cite{llama3modelcard} & 
52.82&
48.42&
48.31&
53.28 &  
48.62& 
32.39&
31.96&
35.59&
38.22& 
34.42

\\
& PMC-LLaMA-13B \cite{wu2024pmc} & 
29.58&
18.04&
17.27&
21.24&  
18.83& 
21.13&
13.92&
15.29&
16.99& 
15.83
 
\\ 
& Meditron-70B \cite{chen2023meditron}  & 
35.21&
24.68&
26.02&
28.96&  
27.19& 
26.76&
24.05&
24.50&
24.32& 
24.59
\\

\midrule
\multirow{4}{*}{\rotatebox{90}{\begin{tabular}[c]{@{}c@{}}  Commer. \\ LLMs \end{tabular}}}

  &  GPT-4o \cite{hurst2024gpt4o}    
  &58.17$_{(1.76)}$ 
  &52.28$_{(1.01)}$ 
  &50.15$_{(0.16)}$ 
  &53.28$_{(0.98)}$ 
  &50.68$_{(0.20)}$
  &31.69$_{(3.80)}$ 
  &35.51$_{(0.88)}$ 
  &39.49$_{(0.49)}$ 
  &39.54$_{(2.02)}$ 
  &38.39$_{(0.79)}$

\\

 &  o3-mini \cite{jaech2024openaio1}   & 58.59$_{(2.41)}$ &
53.61$_{(0.79)}$  &
52.16$_{(0.60)}$  &
52.59$_{(1.02)}$  &
52.55$_{(0.60)}$ &
51.97$_{(3.34)}$ &
59.11$_{(0.99)}$ &
55.54$_{(0.97)}$ &
61.70$_{(4.64)}$ &
55.91$_{(1.02)}$ \\

& o1 \cite{jaech2024openaio1}  & 
61.27&
56.33&
56.59&
55.60& 
56.25& 
60.56&
58.23&
58.46&
61.00& 
58.77

\\

&GPT-4.5 \cite{gpt_4.5} & 
57.04&
55.38&
53.79&
57.53& 
54.22& 
39.44&
37.66&
39.79&
42.47& 
39.04
\\
\midrule

\multirow{4}{*}[-3pt]{\rotatebox{90}{\begin{tabular}[c]{@{}c@{}} AI \\ Agents \end{tabular}}} 

& ReAct-8B \cite{yao2023react}    
& 52.82 & 53.80 & 50.18 & 49.81 &  51.14 & 34.51 & 40.82 & 44.57 & 35.91 & 40.50

\\

& BOLAA-8B  \cite{liu2023bolaa}  
& 55.63 & 56.65  & 55.43 & 52.90 & 52.60 & 40.85 & 47.47 & 50.41 & 46.33 & 46.02

\\ 

& Chameleon-8B  \cite{lu2024chameleon} 
 & 57.04 & 57.91 & 53.56 & 55.21 & 54.30 & 38.73 & 50.32 & 48.42 & 49.81 & 47.89

\\

\cmidrule{2-12}

 &  RiskAgent-8B    & 
\bf 74.09$_{(2.06)}$ &
    \bf 76.84$_{(1.05)}$ &
\bf 78.11$_{(0.72)}$ &
\bf 71.51$_{(0.75)}$ &
\bf 78.34$_{(0.43)}$ &
\bf 77.74$_{(1.58)}$ &
\bf 76.01$_{(0.31)}$ &
\bf 75.61$_{(0.86)}$ &
\bf 70.50$_{(0.93)}$ &
\bf 76.33$_{(0.73)}$
\\

\bottomrule
\end{tabular}
\vspace{-5pt}
\end{table*}

\subsection{Instruction Fine-tuning}

Following the workflow established in Environments 1-5, we construct samples with explicit instructions and well-defined output formats. Each sample includes: (a) patient information, clinical queries, and candidate tools, with the expected output being the selection of the correct tool with justification; (b) patient information and tool schema, with the expected output being correctly formatted parameter extraction; (c) the original results computed by calculators with the sample question, with the expected output being a rephrased answer to the question; (d) outputs from step c along with questions and options, with the expected output being the final selected option; and (e) information from steps a-d, with the expected output being a comprehensive reflection (the instruction template is shown in Figure \ref{fig:prompts}). 
We take the LLaMA-3-8B LLM \cite{llama3modelcard} as our backbone to train our model using parameter-efficient fine-tuning techniques (LoRA) \cite{hu2022lora}, significantly reducing computational requirements while maintaining high performance across medical risk prediction tasks.

It is worth noticing that we only train a single LLaMA-3-8B model for all the three agents across five environments rather than developing separate model for each. This ensures that the model parameters of our system is 8B making it fairly comparable to existing methods and LLMs.

\vspace{-5pt}
\section{Experiments}
In this section, we evaluate the effectiveness of our solution for risk prediction, tool learning, and question answering \footnote{We attach a detailed description of our settings in Appendix~\ref{appen:method}}.

\vspace{-2pt}
\subsection{Evaluation on Risk Prediction}
\paragraph{MedRisk Benchmark}
We build a benchmark MedRisk consisting of diverse risk problems for evaluating LLMs' performance for risk prediction.
We first collect all available evidence-based tools, i.e., clinical calculators, from clinical-standard source MDCalc \cite{elovic2019mdcalc}.
We then review and exclude tools that are not directly predicting disease risks (e.g., those predicting Pregnancy Due Dates and BMI), retaining 387 tools that predict 154 diseases across different scenarios.
We use the APIs provided by MDCalc to generate over 15,000 questions with free-text reference answers by requesting GPT-4o \cite{gpt-4} in Azure to randomly select appropriate tool parameters, 20 quantitative and 20 qualitative questions for each tool.
Next, we manually identify and exclude data samples that are unrelated to risk prediction, obtaining 12,352 risk questions across 154 diseases, 86 symptoms, 50 specialties, and 24 organ systems.
For robust and efficient evaluation, we convert these questions and free-text references to multiple-choice format by combining the correct answer with three wrong choices.
We finally divide the benchmark into MedRisk-Qualitative and -Quantitative, with 6,176 data samples each, according to their question and answer types. For example, the question `What is the estimated risk of postoperative pulmonary complications?' (Answer: High risk) belongs to Qualitative, while the question `What is the Risk Score for Asthma Exacerbation (RSE) for this patient?' (Answer: 30-37\%) belongs to Quantitative.
Figure~\ref{fig:introduction}(b) shows the statistics of the top-30 diseases, top-30 symptoms, top-15 specialties, and top-15 organ systems in the benchmark.

\begin{table*}[t]
\centering
\scriptsize
\caption{Performance of tool learning on the external MEDCALC-BENCH \cite{khandekar2025medcalc} benchmark, which is categorized into Equation-based (Lab, Physical, Date, Dosage) and Rule-based (Risk, Severity, Diagnosis) tools.
}
\label{tab:medical_calc_cot_results}
\setlength{\tabcolsep}{3pt}

\begin{tabular}{clccccccccc}
\toprule
\multirow{2}{*}[-3pt]{\textbf{Types}} &  \multirow{2}{*}[-3pt]{\textbf{Methods}}   & \multicolumn{4}{c}{\textbf{Equation-based Diagnosis}} & \multicolumn{3}{c}{\textbf{Rule-based Diagnosis}} & \multirow{2}{*}[-3pt]{\textbf{Overall}}    \\
\cmidrule(lr){3-6} \cmidrule(lr){7-9}
 & &  \includegraphics[width=0.3cm]{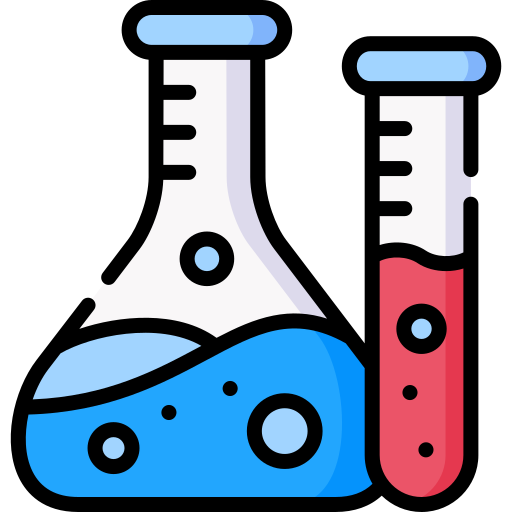} Lab   & \includegraphics[width=0.3cm]{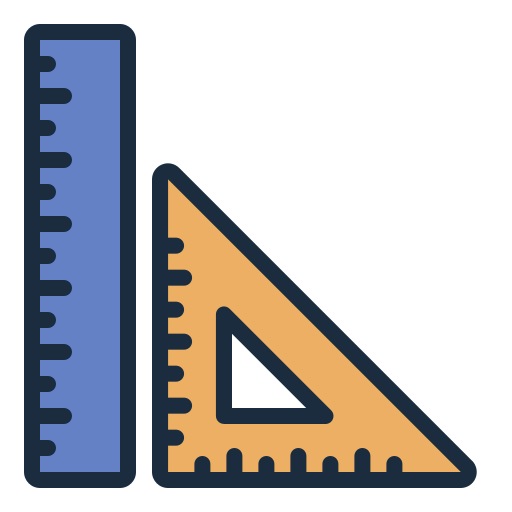} Physical & \includegraphics[width=0.3cm]{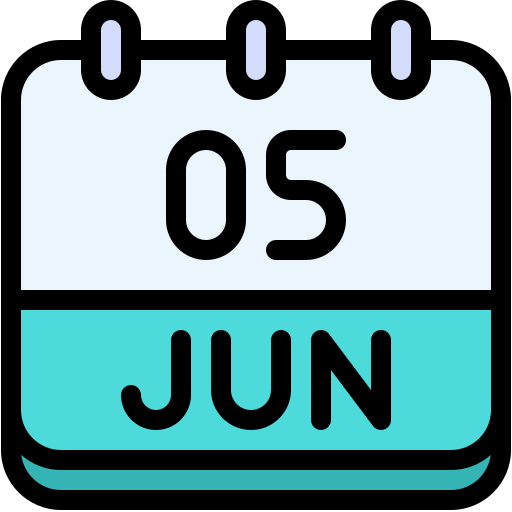} Date & \includegraphics[width=0.3cm]{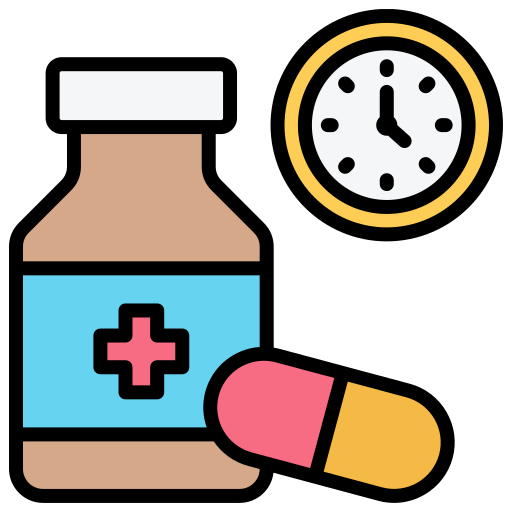} Dosage & \includegraphics[width=0.3cm]{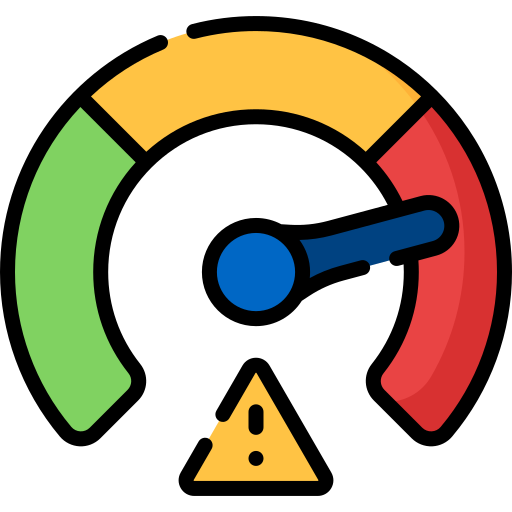} Risk & \includegraphics[width=0.3cm]{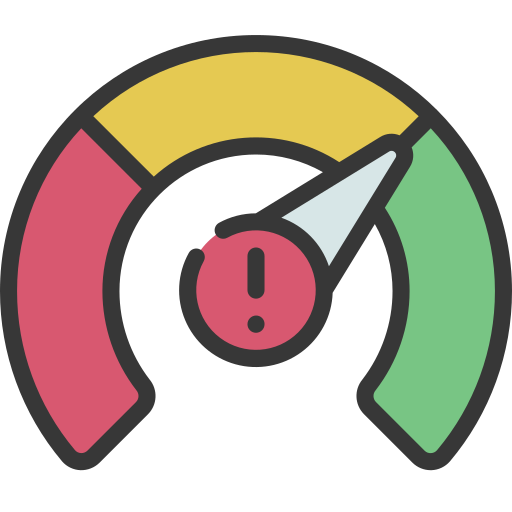} Severity & \includegraphics[width=0.3cm]{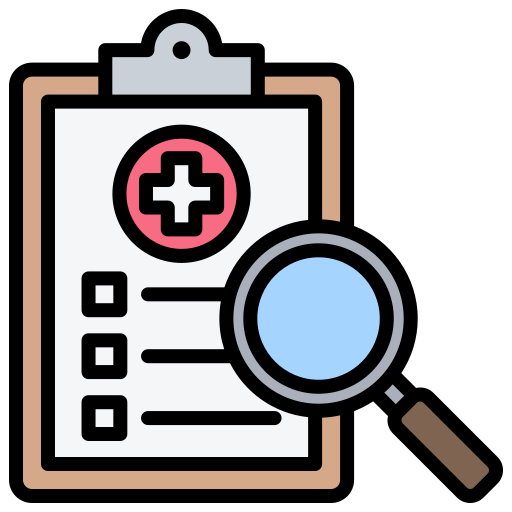} Diagnosis & \\
\midrule
\multirow{7}{*}{\rotatebox{90}{\begin{tabular}[c]{@{}c@{}} Public \\ LLMs \end{tabular}}} 
& PMC-LLaMA-13B \cite{wu2024pmc}   &
0.00 & 0.00 & 0.00 & 0.00 & 0.00 & 0.00 & 0.00 & 0.00 \\

& Meditron-70B \cite{chen2023meditron}    &
0.00 & 0.00 & 3.33 & 0.00 & 0.00 & 0.00 & 3.33 & 0.38 \\

&  Mistral-7B \cite{jiang2023mistral}  &
10.09 & 14.58 & 1.67 & 0.00 & 9.58 & 7.50 & 25.00 & 10.79 \\

&LLaMA-3-8B \cite{llama3modelcard}   &
16.51 & 25.00 & 1.67 & 7.50 & 11.25 & 13.75 & 26.67 & 16.43 \\

&Mixtral-8x7B \cite{jiang2024mixtral}   &
\textbf{22.63} & 40.00 & 6.67 & 17.50 & 11.25 & \textbf{21.25} & 15.00 & 22.35 \\

 \cmidrule(lr){2-10}
& RiskAgent-8B (Ours)   &
21.41 & \textbf{47.08} & \textbf{10.00} & \textbf{50.00} & \textbf{19.58} & 12.50 & \textbf{33.33} & \textbf{27.32} \\ 
\midrule

\multirow{3}{*}{\rotatebox{90}{\begin{tabular}[c]{@{}c@{}} Commercial \\ LLMs \end{tabular}}}  & GPT-3.5 \cite{openai2022chatgpt}  &
20.49 & 45.00 & 11.67 & 17.50 & 13.33 & 10.00 & 31.67 & 23.69 \\

& GPT-4 \cite{gpt-4}   &
26.30 & 71.25 & \textbf{48.33} & 40.00 & 27.50 & 15.00 & 28.33 & 37.92 \\

& GPT-4o \cite{hurst2024gpt4o}  &
40.98 & 82.50 & 36.67 & 47.50 & 32.50 & 22.50 & 23.33 & 46.13 \\

 \cmidrule(lr){2-10}
 
&RiskAgent-GPT-4o (Ours)  &
\textbf{65.75} & \textbf{97.08} & \textbf{48.33} & \textbf{95.00} & \textbf{52.92} & \textbf{37.50} & \textbf{61.67} & \textbf{67.71} \\

\bottomrule
\end{tabular}
\vspace{-5pt}
\end{table*}

\begin{table*}[t]
\centering
\scriptsize
\caption{
Performance of medical reasoning and question answering on three widely used benchmarks MedQA, MedMCQA, and MMLU.
}
\label{tab:qa}
\setlength{\tabcolsep}{3pt}

\begin{tabular}{lcccccc}
\toprule
 \multirow{2}{*}[-3pt]{\textbf{Methods}}   & \multirow{2}{*}[-3pt]{\textbf{MedQA}}  & \multirow{2}{*}[-3pt]{\textbf{MedMCQA}}   & \multicolumn{4}{c}{\textbf{MMLU}}    \\
  \cmidrule(lr){4-7}
&   &  & Clinical Knowledge & College Medicine & Medical Genetics & Professional Medicine \\
\midrule

 MedGemma \cite{sellergren2025medgemma} & 83.6 & 65.3 & 80.0& 	76.9& 	83.0& 	68.4 \\

 GPT-4o \cite{hurst2024gpt4o} & 86.0  & 77.0  & 88.3 & 85.5 & 	\textbf{96.0} & 	93.8 \\
 
Claude-3.7-Sonnet \cite{Claude37Sonnet} & 78.0 & 72.1 &85.7 &	78.0&	94.0&	89.7
  
  \\

Claude-4-Sonnet \cite{Claude4Sonnet} & 78.6 & 74.5 &86.8&	78.6&	95.0&	91.5 
  \\

 Claude-4-Opus \cite{Claude4Opus} & 79.3 & 77.8 &\textbf{91.7}&	78.0&	\textbf{96.0}&	91.9 
  \\

 \cmidrule(lr){2-7}
 RiskAgent-GPT-4o (Ours) &
\textbf{87.8} &\textbf{80.8}  &89.4 	&\textbf{86.6}&\textbf{96.0} 	&\textbf{95.2 }
\\

\bottomrule
\end{tabular}
\vspace{-5pt}
\end{table*}

\begin{table*}[t]
\centering
\scriptsize
\caption{
Ablation study of our RiskAgent, which includes four main components: Tool Library, Decider, Executor, and Reviewer.
We take LLaMA-3-8B as the basic LLM and use accuracy to report our results.
}
\label{tab:ablation}
\setlength{\tabcolsep}{3pt} 
\begin{tabular}{@{}lcccccccccc@{}}
\toprule
\multirow{3}{*}[-3pt]{Settings} &  \multirow{3}{*}[-3pt]{Tool Library} &  \multirow{3}{*}[-3pt]{Decider} &  \multirow{3}{*}[-3pt]{Executor}  & \multirow{3}{*}[-3pt]{Reviewer} & \multicolumn{3}{c}{\textbf{MedRisk
Qualitative}} & \multicolumn{3}{c}{\textbf{MedRisk
Quantitative}} 

\\ \cmidrule(lr){6-8} \cmidrule(lr){9-11}

&&&&&
 \begin{tabular}[c]{@{}c@{}} Tool \\ Selection  
\end{tabular}   &  \begin{tabular}[c]{@{}c@{}}Parameter \\ Parsing
\end{tabular}     & \begin{tabular}[c]{@{}c@{}} Overall \\ Performance
\end{tabular}   
& \begin{tabular}[c]{@{}c@{}} Tool \\ Selection  
\end{tabular}   &  \begin{tabular}[c]{@{}c@{}}Parameter \\ Parsing
\end{tabular}     & \begin{tabular}[c]{@{}c@{}} Overall\\Performance
\end{tabular}   
\\ 
\midrule

 LLaMA-3-70B \cite{llama3modelcard}   & -   & -  & - & - & -  & - &48.62&-&-&34.42
 
\\
 GPT-4o \cite{hurst2024gpt4o}    
  & -   & -  & - & - & -  & -  &50.68&- & - & 38.39

 \\

\midrule

Basic LLM    & -   & -  & - & - & -  & - & 43.75 & - & - & 32.06
\\ 
 \cmidrule(lr){2-11}
(a)  & $\checkmark$ & -   &-&-& 82.06 &  35.06 & 57.95 & 78.49	& 32.71 & 46.59
\\
(b)  & $\checkmark$  & $\checkmark$ & $\checkmark$ & - & 84.50 & 62.17 & 72.00 & 88.15	& 62.91 & 65.58
 \\
 \cmidrule(lr){2-11}
RiskAgent-8B & $\checkmark$ & $\checkmark$ & $\checkmark$  & $\checkmark$  & \bf 92.94 & \bf 68.02 & \bf 78.34 & \bf 97.89 &\bf  71.19&\bf 76.33
\\ 
\bottomrule
\end{tabular}
\vspace{-10pt}
\end{table*}

\paragraph{Results}

In Table~\ref{tab:results}, we report different methods' performance on the most common cases across disease, symptom, specialty, and organ system and their overall performance.
RiskAgent not only outperforms different types of LLM in all cases, respectively, but also achieves the highest overall accuracy of 78.34\% and 76.33\% in MedRisk-Qualitative and MedRisk-Quantitative, respectively, doubling the 38.39\% accuracy of GPT-4o. It indicates the robustness of our method across different diseases, symptoms, specialties, organ system, and question types, providing a more comprehensive risk prediction solution than previous methods.
In detail, compared with large commercial LLMs (e.g., GPT-4o, GPT-o1, and GPT-o3-mini) and open-source LLMs (e.g., Meditron-70B and LLaMA-3-70B), our RiskAgent outperforms them by $>$20.0\% accuracy, with much (at least 10 times) fewer parameters. This is desirable in resource-limited medical settings.
Our method also achieves the best results among AI agents, with accuracy surpassing previous methods by up to 35.83\% overall accuracy.

To clearly understand the effectiveness of our approach, we visualize the results on MedRisk-Quantitative in Figure \ref{fig:introduction}(c), which shows that our RiskAgent outperforms previous methods in all cases.
These encouraging results demonstrate that our RiskAgent achieves more precise and robust decision support than existing advanced LLMs for complex medical tasks. Importantly, our method uses evidence-based tools for prediction and traces the information source (i.e., evidence) behind our system's decisions, making the results more transparent and reliable than existing approaches.

\subsection{Evaluation on Tool Learning}
We further evaluate our method's tool learning ability on an external benchmark MEDCALC-BENCH \cite{khandekar2025medcalc}, which requires the models to use medical calculations for decision support.
Note that our RiskAgent is designed specifically for risk prediction; thus, this external benchmark evaluates our methods' tool learning ability to new tasks and domains.
For evaluation, we follow previous work \cite{khandekar2025medcalc} to employ zero-shot chain-of-thought (CoT) prompting to report the results. We adopt different backbones, i.e., LLaMA-3-8B \cite{llama3modelcard} and GPT-4o \cite{hurst2024gpt4o}, to implement two variants of our method: RiskAgent-8B and RiskAgent-GPT-4o. 

Table~\ref{tab:medical_calc_cot_results} 
clearly show that: (i) The RiskAgent-8B model, with only 8B parameters, surpasses all public LLMs, including the larger LLMs Meditron-70B \cite{chen2023meditron} and Mixtral-8x7B \cite{jiang2024mixtral} in major cases.
(ii) The RiskAgent-GPT-4o model significantly outperforms all baseline models with an overall accuracy of 67.71\%, representing a substantial improvement over the standard GPT-4 and GPT-4o. Most notably, our model exhibits exceptional performance in rule-based diagnosis tasks, achieving 61.67\% accuracy in the Diagnosis category - more than double the performance of all baseline methods.
Meanwhile, the performance improvements are particularly striking in calculation-heavy categories, with RiskAgent-GPT-4o achieving 97.08\% on Physical assessments and 95.00\% on Dosage calculations - areas where precise quantitative reasoning is critical for patient safety. 
It can also be verified by our RiskAgent-8B model, which, outperforms all public and commercial LLMs in dosage calculations.

\subsection{Evaluation on Question Answering}
We further adopt three widely used benchmarks, MedQA \cite{jin2021disease}, MedMCQA \cite{pal2022medmcqa}, and MMLU \cite{hendrycks2020measuring}, to evaluate the performance of our method in general medical reasoning and question answering.
As shown in Table~\ref{tab:qa}, RiskAgent achieves the best results on most datasets.
This demonstrates that although our method is designed for risk prediction, it exhibits strong generalization capabilities across diverse diagnostic tasks in previously unseen clinical scenarios and maintains robust medical reasoning performance on general questions.

Overall, these encouraging results further underscore the effectiveness and robustness of our approach and its potential to provide reliable decision support across various medical contexts by synergizing language models with validated tools for evidence-based medical recommendations.

\vspace{-5pt}
\section{Analysis}

\vspace{-1pt}
\subsection{Ablation Study}
\vspace{-1pt}

Our method introduces four main components, i.e., the Tool Library in the environment, the Decider, the Executor, and the Reviewer, to achieve superior performance.
(i) As shown in Table~\ref{tab:ablation},
all components contribute to improved performance, proving the effectiveness of each component.
(ii) By comparing Basic LLM and Setting (a), which introduces the tool library to enable the basic LLM to use the evidence-based tool for risk prediction, we observe that Setting (a) significantly boosts the performance of the basic LLM by more than 14\% accuracy, surpassing larger models like LLaMA-3-70B and GPT-4o. This proves the motivation and effectiveness of leveraging evidence-based tools to achieve improved performance.
(iii) Considering that collaborating with tools involves both tool selection and parameter parsing for tool execution, we further report their accuracy.
Comparing Setting (a) and Setting (b), we observe that introducing LLM agents can significantly boost performance—improving tool selection accuracy by 9.66\%, parameter parsing accuracy by 30.20\%, and overall task performance by 18.99\%.
This proves the effectiveness of our method in separating the collaboration process into tool selection and parameter parsing through the designed Decider and Executor. 
By distributing responsibilities, our approach reduces the model’s learning burden, enabling accurate and efficient synergy with medical tools.
(iv) Finally, we notice that the Reviewer further improves overall performance, showing the effectiveness of reflecting on the analysis process and thereby correcting potential errors.

\begin{figure}[t]
\centering
\includegraphics[
  width=0.85\linewidth,
  trim=13cm -0.3cm 0 0,
  clip
]{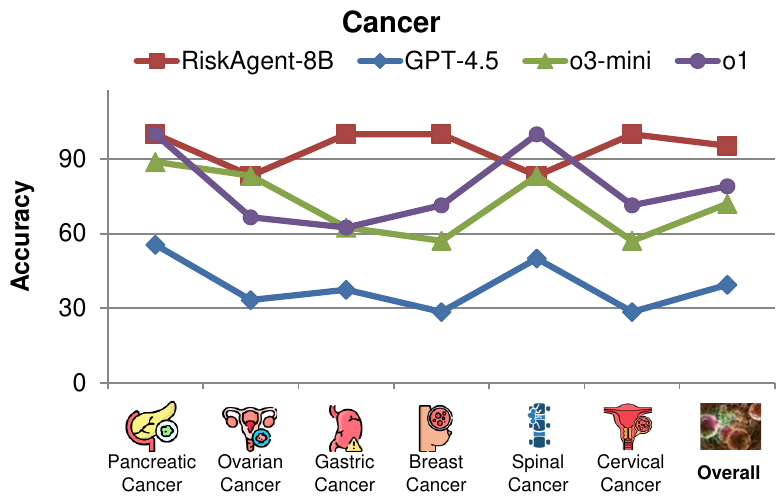}
\\
\includegraphics[
  width=0.85\linewidth,
  trim=0 0 13cm 0,
  clip
]{cancer+rare.pdf}
\vspace{-5pt}
\caption{
We evaluate the performance of models on five rare diseases (left) and six types of cancer (right).
SM: Systemic Mastocytosis;
PM: Primary Myelofibrosis;
CML: Chronic Myelogenous Leukemia;
IPF: Idiopathic Pulmonary Fibrosis.
}
\label{fig:rare_cancer}
\vspace{-10pt}
\end{figure}

\begin{figure*}[t]
\centering
\includegraphics[width=0.7\linewidth]{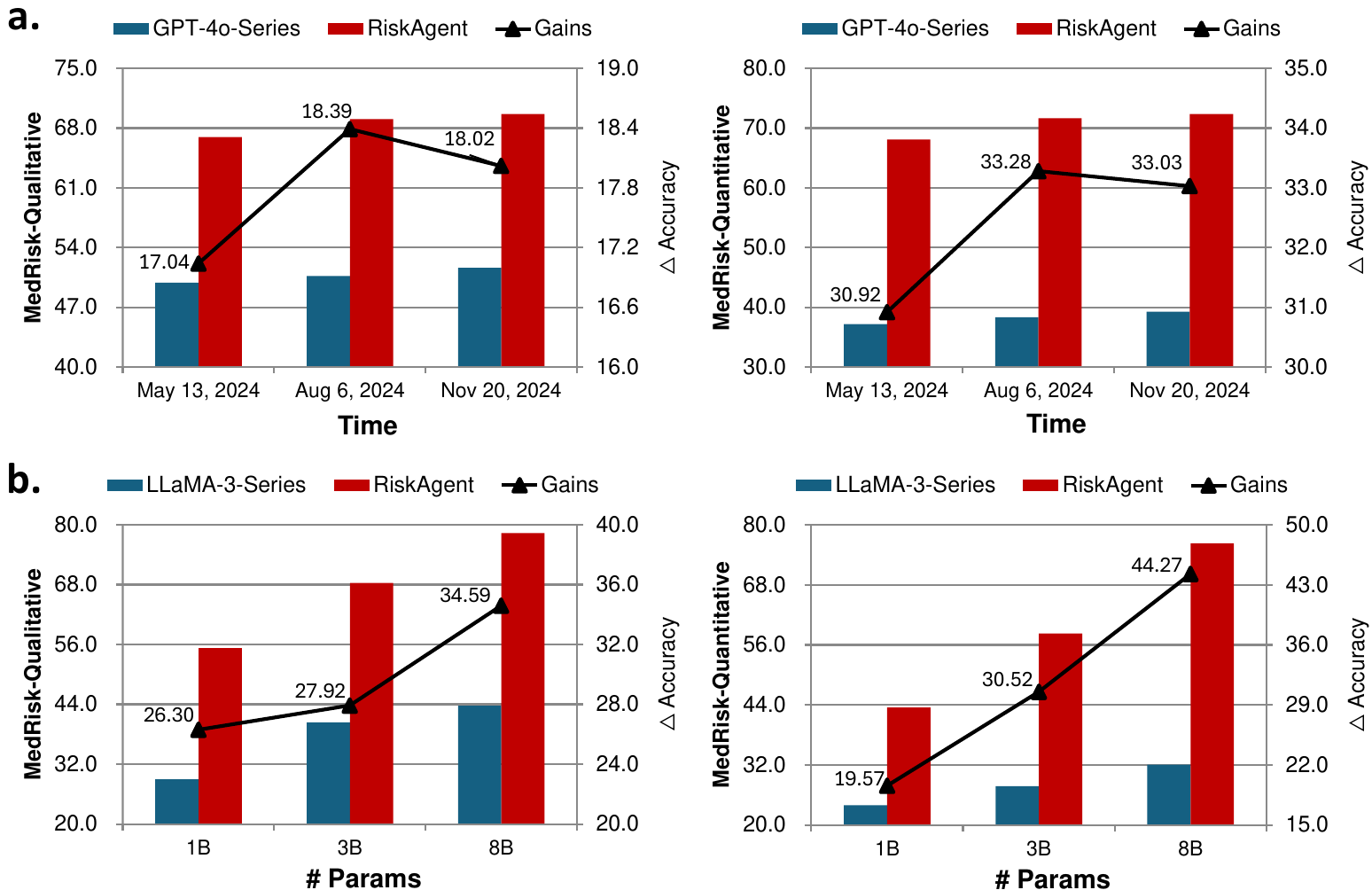}
\vspace{-5pt}
\caption{The generalization ability of our method: We report the overall accuracy of the basic LLMs (blue bars) and the basic LLMs enhanced using our method (red bars). We evaluate \textbf{a.} different variants of the GPT-4o-series LLMs developed at different times and \textbf{b.} LLaMA-3-series LLMs with varying numbers of model parameters. 
The polyline and the right y-axis show the improvements in different variants. 
We can see that the more advanced (\textbf{a.}) and the larger (\textbf{b.}) the basic LLM, the greater the improvements. 
}
\label{fig:generalization}
\vspace{-5pt}
\end{figure*}

\subsection{Effect of Rare Diseases and Cancer}
We further conduct experiments on rare diseases and cancer to evaluate our method's robustness in solving diverse complex diseases.
For rare (but important) diseases, we follow two clinical-standard rare disease databases, i.e., European Orphanet (\url{https://www.orpha.net/}) and the US National Organization for Rare Disorders (NORD) (\url{https://rarediseases.org/}), to choose five rare diseases for evaluation. The upper part of Figure~\ref{fig:rare_cancer} shows that RiskAgent achieves the best results. 
Due to the scarcity of rare disease data, previous LLMs struggle to learn sufficient knowledge during their pretraining, making it difficult for them to provide accurate analysis and predictions for rare diseases.
Fortunately, our method collaborates with specialized tools for these rare diseases to leverage their evidence-based knowledge and advantages, achieving significantly improved performance. 
For cancer, as shown in the bottom part of Figure~\ref{fig:rare_cancer}, we can see that our method consistently achieves competitive results with different cancers.
It shows the potential of our solution, with only 8B parameters, to provide more accurate and robust risk prediction for complex diseases compared to state-of-the-art LLMs.

\subsection{Generalization Analysis}
 
We provide a solution that synergizes LLMs with validated tools to produce accurate, reliable medical results.
Thus, our method is agnostic to the underlying LLM, making it adaptable to various LLMs and helping them produce accurate and evidence-based predictions.
To this end, we further explore our solution's generalizability and applicability as LLMs continue to evolve rapidly over time and as the number of parameters increases.
We select the GPT-4o-series developed at different times (\textit{gpt-4o-2024-05-13}, \textit{gpt-4o-2024-08-06}, and \textit{gpt-4o-2024-11-20}) and LLaMA-series (\textit{LLaMA-3-1B}, \textit{LLaMA-3-3B}, and \textit{LLaMA-3-8B}), which share the architecture but differ in model size. 

We compare the performance of (i) the basic LLM and (ii) the basic LLM enhanced using our method. Figure~\ref{fig:generalization} shows that our method provides significant performance improvements for all basic LLMs (accuracy of 17.04\%$\sim$44.27\%) on both MedRisk-Qualitative and MedRisk-Quantitative benchmarks.
This indicates that if downstream applications do not involve strict privacy regulations, our method can significantly improve the performance of large advanced LLMs, enabling them to solve complex medical problems.
Even when privacy concerns are involved, our method can still significantly boost the results of LLaMA-3-series public LLMs across different parameters.
For example, our method improves accuracy by 19.57\%$\sim$30.52\% for the 1B and 3B LLMs, enabling them, with only 1B and 3B model parameters, to outperform the commercial GPT-4o basic LLMs.
Such encouraging improvements in lightweight LLMs are useful in resource-limited medical institutions.
Meanwhile, we notice that as model sizes increase, the improvements brought by our method become greater. Overall, the improved results across different GPT-4o-series and LLaMA-3-series LLMs show the strong potential of our approach, enabling large and advanced LLMs to deliver accurate medical data analysis and evidence-based predictions. As LLMs continue to evolve rapidly, our method can further enhance their effectiveness in complex and critical tasks.

\section{Conclusion}
In this paper, we present RiskAgent to achieve faithful and evidence-based medical recommendations. By grounding recommendations in valiated tools, RiskAgent offers a resource-efficient, transparent, and trustworthy solution for complex clinical decision-making.
Extensive experiments demonstrate that RiskAgent, using only 8 billion parameters, achieves competitive performance with state-of-the-art models in risk prediction, tool learning, and medical reasoning.

The encouraging results highlight the strong potential of our approach across diverse medical domains. For example, instead of extensively fine-tuning LLMs for different tasks, as in common practice, our method collaborates with and leverages validated tools, not only producing trustworthy results but also reducing resource costs, thereby making LLMs more accessible for resource-limited medical applications.

\section*{Impact Statement}
This work advances clinical AI by providing potentially expert-level risk prediction for resource-limited institutions, as our 8B model runs effectively on accessible hardware. By leveraging validated medical tools, RiskAgent mitigates the hallucination issues common in general LLMs, aligning AI outputs with evidence-based medicine. While this approach enhances transparency, real-world deployment must strictly adhere to privacy regulations regarding patient data. Furthermore, RiskAgent is designed as a decision support tool; clinician oversight remains essential to ensure patient safety and validate tool applicability.

Our study was conducted on retrospective, de-identified data, in which all Protected Health Information (PHI)—such as patient name, sex, gender, and date of birth—was officially de-identified in compliance with the Health Insurance Portability and Accountability Act (HIPAA) standards.

\section*{Acknowledgements}
DAC was funded by an NIHR Research Professorship; a Royal Academy of Engineering Research Chair; and the InnoHK Hong Kong Centre for Cerebro-cardiovascular Engineering (COCHE); and was supported by the National Institute for Health Research (NIHR) Oxford Biomedical Research Centre (BRC) and the Pandemic Sciences Institute at the University of Oxford. FL was funded by the Clarendon Fund and Magdalen Graduate Scholarship. HZ was funded by the Clarendon Fund, the Department of Engineering Science Studentship, and the Frederick Brodckhues Scholarship.

The Applied Digital Health (ADH) group at the Nuffield Department of Primary Care Health Sciences is supported by the National Institute for Health and Care Research (NIHR) Applied Research Collaboration Oxford and Thames Valley at Oxford Health NHS Foundation Trust. The views expressed are those of the author(s) and not necessarily those of the NHS, the NIHR or the Department of Health and Social Care.

\bibliography{refs,refs2,refs3}

\begin{thebibliography}{54}
\providecommand{\natexlab}[1]{#1}
\providecommand{\url}[1]{\texttt{#1}}
\expandafter\ifx\csname urlstyle\endcsname\relax
  \providecommand{\doi}[1]{doi: #1}\else
  \providecommand{\doi}{doi: \begingroup \urlstyle{rm}\Url}\fi

\bibitem[tex(2022)]{text-embedding-ada-002}
New and improved embedding model.
\newblock \url{https://openai.com/index/new-and-improved-embedding-model/}, 2022.

\bibitem[gpt(2025{\natexlab{a}})]{gpt2025o3}
Openai o3-mini.
\newblock \url{https://openai.com/index/openai-o3-mini/}, 2025{\natexlab{a}}.

\bibitem[gpt(2025{\natexlab{b}})]{gpt_4.5}
Openai gpt-4.5 system card.
\newblock \url{https://openai.com/index/gpt-4-5-system-card/}, 2025{\natexlab{b}}.

\bibitem[AI@Meta(2024)]{llama3modelcard}
AI@Meta.
\newblock Llama 3 model card, 2024.
\newblock URL \url{https://github.com/meta-llama/llama3/blob/main/MODEL_CARD.md}.

\bibitem[Anthropic(2024{\natexlab{a}})]{Claude37Sonnet}
Anthropic.
\newblock Claude-3.7-sonnet.
\newblock \url{https://www.anthropic.com/claude}, 2024{\natexlab{a}}.

\bibitem[Anthropic(2024{\natexlab{b}})]{Claude4Opus}
Anthropic.
\newblock Claude-4-opus.
\newblock \url{https://www.anthropic.com/claude}, 2024{\natexlab{b}}.

\bibitem[Anthropic(2024{\natexlab{c}})]{Claude4Sonnet}
Anthropic.
\newblock Claude-4-sonnet.
\newblock \url{https://www.anthropic.com/claude}, 2024{\natexlab{c}}.

\bibitem[Chen et~al.(2023)Chen, Cano, Romanou, Bonnet, Matoba, Salvi, Pagliardini, Fan, K{\"o}pf, Mohtashami, et~al.]{chen2023meditron}
Chen, Z., Cano, A.~H., Romanou, A., Bonnet, A., Matoba, K., Salvi, F., Pagliardini, M., Fan, S., K{\"o}pf, A., Mohtashami, A., et~al.
\newblock Meditron-70b: Scaling medical pretraining for large language models.
\newblock \emph{arXiv preprint arXiv:2311.16079}, 2023.

\bibitem[Chowdhery et~al.(2022)Chowdhery, Narang, Devlin, Bosma, Mishra, Roberts, Barham, Chung, Sutton, Gehrmann, et~al.]{chowdhery2022palm}
Chowdhery, A., Narang, S., Devlin, J., Bosma, M., Mishra, G., Roberts, A., Barham, P., Chung, H.~W., Sutton, C., Gehrmann, S., et~al.
\newblock Palm: Scaling language modeling with pathways.
\newblock \emph{arXiv preprint arXiv:2204.02311}, 2022.

\bibitem[Damen et~al.(2016)Damen, Hooft, Schuit, Debray, Collins, Tzoulaki, Lassale, Siontis, Chiocchia, Roberts, et~al.]{damen2016prediction}
Damen, J.~A., Hooft, L., Schuit, E., Debray, T.~P., Collins, G.~S., Tzoulaki, I., Lassale, C.~M., Siontis, G.~C., Chiocchia, V., Roberts, C., et~al.
\newblock Prediction models for cardiovascular disease risk in the general population: systematic review.
\newblock \emph{bmj}, 353, 2016.

\bibitem[Elovic \& Pourmand(2019)Elovic and Pourmand]{elovic2019mdcalc}
Elovic, A. and Pourmand, A.
\newblock Mdcalc medical calculator app review.
\newblock \emph{Journal of digital imaging}, 32:\penalty0 682--684, 2019.

\bibitem[Ferber et~al.(2024)Ferber, Wiest, W{\"o}lflein, Ebert, Beutel, Eckardt, Truhn, Springfeld, J{\"a}ger, and Kather]{ferber2024gpt}
Ferber, D., Wiest, I.~C., W{\"o}lflein, G., Ebert, M.~P., Beutel, G., Eckardt, J.-N., Truhn, D., Springfeld, C., J{\"a}ger, D., and Kather, J.~N.
\newblock Gpt-4 for information retrieval and comparison of medical oncology guidelines.
\newblock \emph{NEJM AI}, 1\penalty0 (6):\penalty0 AIcs2300235, 2024.

\bibitem[Google(2023)]{google2023bard}
Google.
\newblock Bard: A generative artificial intelligence chatbot.
\newblock \url{https://gemini.google.com}, 2023.

\bibitem[Grondman et~al.(2012)Grondman, Busoniu, Lopes, and Babuska]{grondman2012survey}
Grondman, I., Busoniu, L., Lopes, G.~A., and Babuska, R.
\newblock A survey of actor-critic reinforcement learning: Standard and natural policy gradients.
\newblock \emph{IEEE Transactions on Systems, Man, and Cybernetics, part C (applications and reviews)}, 42\penalty0 (6):\penalty0 1291--1307, 2012.

\bibitem[Han et~al.(2023)Han, Adams, Papaioannou, Grundmann, Oberhauser, L{\"o}ser, Truhn, and Bressem]{han2023medalpaca}
Han, T., Adams, L.~C., Papaioannou, J.-M., Grundmann, P., Oberhauser, T., L{\"o}ser, A., Truhn, D., and Bressem, K.~K.
\newblock Medalpaca--an open-source collection of medical conversational ai models and training data.
\newblock \emph{arXiv preprint arXiv:2304.08247}, 2023.

\bibitem[Hendrycks et~al.(2020)Hendrycks, Burns, Basart, Zou, Mazeika, Song, and Steinhardt]{hendrycks2020measuring}
Hendrycks, D., Burns, C., Basart, S., Zou, A., Mazeika, M., Song, D., and Steinhardt, J.
\newblock Measuring massive multitask language understanding.
\newblock \emph{arXiv preprint arXiv:2009.03300}, 2020.

\bibitem[Hu et~al.(2022)Hu, Shen, Wallis, Allen-Zhu, Li, Wang, Wang, Chen, et~al.]{hu2022lora}
Hu, E.~J., Shen, Y., Wallis, P., Allen-Zhu, Z., Li, Y., Wang, S., Wang, L., Chen, W., et~al.
\newblock Lora: Low-rank adaptation of large language models.
\newblock \emph{ICLR}, 1\penalty0 (2):\penalty0 3, 2022.

\bibitem[Hurst et~al.(2024)Hurst, Lerer, Goucher, Perelman, Ramesh, Clark, Ostrow, Welihinda, Hayes, Radford, et~al.]{hurst2024gpt4o}
Hurst, A., Lerer, A., Goucher, A.~P., Perelman, A., Ramesh, A., Clark, A., Ostrow, A., Welihinda, A., Hayes, A., Radford, A., et~al.
\newblock Gpt-4o system card.
\newblock \emph{arXiv preprint arXiv:2410.21276}, 2024.

\bibitem[Jaech et~al.(2024)Jaech, Kalai, Lerer, Richardson, El-Kishky, Low, Helyar, Madry, Beutel, Carney, et~al.]{jaech2024openaio1}
Jaech, A., Kalai, A., Lerer, A., Richardson, A., El-Kishky, A., Low, A., Helyar, A., Madry, A., Beutel, A., Carney, A., et~al.
\newblock Openai o1 system card.
\newblock \emph{arXiv preprint arXiv:2412.16720}, 2024.

\bibitem[Ji et~al.(2023)Ji, Lee, Frieske, Yu, Su, Xu, Ishii, Bang, Madotto, and Fung]{Hallu-survey}
Ji, Z., Lee, N., Frieske, R., Yu, T., Su, D., Xu, Y., Ishii, E., Bang, Y.~J., Madotto, A., and Fung, P.
\newblock Survey of hallucination in natural language generation.
\newblock \emph{ACM Computing Surveys}, 55\penalty0 (12):\penalty0 1--38, 2023.

\bibitem[Jiang et~al.(2023)Jiang, Sablayrolles, Mensch, Bamford, Chaplot, Casas, Bressand, Lengyel, Lample, Saulnier, et~al.]{jiang2023mistral}
Jiang, A.~Q., Sablayrolles, A., Mensch, A., Bamford, C., Chaplot, D.~S., Casas, D. d.~l., Bressand, F., Lengyel, G., Lample, G., Saulnier, L., et~al.
\newblock Mistral {7B}.
\newblock \emph{arXiv preprint arXiv:2310.06825}, 2023.

\bibitem[Jiang et~al.(2024)Jiang, Sablayrolles, Roux, Mensch, Savary, Bamford, Chaplot, Casas, Hanna, Bressand, et~al.]{jiang2024mixtral}
Jiang, A.~Q., Sablayrolles, A., Roux, A., Mensch, A., Savary, B., Bamford, C., Chaplot, D.~S., Casas, D. d.~l., Hanna, E.~B., Bressand, F., et~al.
\newblock Mixtral of experts.
\newblock \emph{arXiv preprint arXiv:2401.04088}, 2024.

\bibitem[Jin et~al.(2021)Jin, Pan, Oufattole, Weng, Fang, and Szolovits]{jin2021disease}
Jin, D., Pan, E., Oufattole, N., Weng, W.-H., Fang, H., and Szolovits, P.
\newblock What disease does this patient have? a large-scale open domain question answering dataset from medical exams.
\newblock \emph{Applied Sciences}, 11\penalty0 (14):\penalty0 6421, 2021.

\bibitem[Jin et~al.(2024)Jin, Wang, Yang, Zhu, Wright, Huang, Wilbur, He, Taylor, Chen, et~al.]{jin2024agentmd}
Jin, Q., Wang, Z., Yang, Y., Zhu, Q., Wright, D., Huang, T., Wilbur, W.~J., He, Z., Taylor, A., Chen, Q., et~al.
\newblock Agentmd: Empowering language agents for risk prediction with large-scale clinical tool learning.
\newblock \emph{arXiv preprint arXiv:2402.13225}, 2024.

\bibitem[Khandekar et~al.(2025)Khandekar, Jin, Xiong, Dunn, Applebaum, Anwar, Sarfo-Gyamfi, Safranek, Anwar, Zhang, et~al.]{khandekar2025medcalc}
Khandekar, N., Jin, Q., Xiong, G., Dunn, S., Applebaum, S., Anwar, Z., Sarfo-Gyamfi, M., Safranek, C., Anwar, A., Zhang, A., et~al.
\newblock Medcalc-bench: Evaluating large language models for medical calculations.
\newblock \emph{Advances in Neural Information Processing Systems}, 37:\penalty0 84730--84745, 2025.

\bibitem[Kim et~al.(2024)Kim, Park, Jeong, Chan, Xu, McDuff, Lee, Ghassemi, Breazeal, and Park]{kim2024mdagents}
Kim, Y., Park, C., Jeong, H., Chan, Y.~S., Xu, X., McDuff, D., Lee, H., Ghassemi, M., Breazeal, C., and Park, H.~W.
\newblock Mdagents: An adaptive collaboration of llms for medical decision-making.
\newblock \emph{Advances in Neural Information Processing Systems}, 37:\penalty0 79410--79452, 2024.

\bibitem[Kitamura(2023)]{kitamura2023chatgpt}
Kitamura, F.~C.
\newblock Chatgpt is shaping the future of medical writing but still requires human judgment.
\newblock \emph{Radiology}, pp.\  230171, 2023.

\bibitem[Liu et~al.(2024)Liu, Li, Zhou, Yin, Yang, Tang, Luo, Zeng, Jiang, Gao, Nigam, Nag, Yin, Hua, Zhou, Rohanian, Thakur, Clifton, and Clifton]{liu2024large}
Liu, F., Li, Z., Zhou, H., Yin, Q., Yang, J., Tang, X., Luo, C., Zeng, M., Jiang, H., Gao, Y., Nigam, P., Nag, S., Yin, B., Hua, Y., Zhou, X., Rohanian, O., Thakur, A., Clifton, L., and Clifton, D.
\newblock Large language models are poor clinical decision-makers: A comprehensive benchmark.
\newblock In \emph{{Proceedings of the 2024 Conference on Empirical Methods in Natural Language Processing }}, 2024.

\bibitem[Liu et~al.(2023)Liu, Yao, Zhang, Xue, Heinecke, Murthy, Feng, Chen, Niebles, Arpit, et~al.]{liu2023bolaa}
Liu, Z., Yao, W., Zhang, J., Xue, L., Heinecke, S., Murthy, R., Feng, Y., Chen, Z., Niebles, J.~C., Arpit, D., et~al.
\newblock Bolaa: Benchmarking and orchestrating llm-augmented autonomous agents.
\newblock \emph{arXiv preprint arXiv:2308.05960}, 2023.

\bibitem[Loshchilov \& Hutter(2019)Loshchilov and Hutter]{loshchilov2019adamW}
Loshchilov, I. and Hutter, F.
\newblock Decoupled weight decay regularization.
\newblock In \emph{International Conference on Learning Representations}, 2019.

\bibitem[Lu et~al.(2024)Lu, Peng, Cheng, Galley, Chang, Wu, Zhu, and Gao]{lu2024chameleon}
Lu, P., Peng, B., Cheng, H., Galley, M., Chang, K.-W., Wu, Y.~N., Zhu, S.-C., and Gao, J.
\newblock Chameleon: Plug-and-play compositional reasoning with large language models.
\newblock \emph{Advances in Neural Information Processing Systems}, 36, 2024.

\bibitem[Nori et~al.(2023)Nori, Lee, Zhang, Carignan, Edgar, Fusi, King, Larson, Li, Liu, et~al.]{nori2023medprompt}
Nori, H., Lee, Y.~T., Zhang, S., Carignan, D., Edgar, R., Fusi, N., King, N., Larson, J., Li, Y., Liu, W., et~al.
\newblock Can generalist foundation models outcompete special-purpose tuning? case study in medicine.
\newblock \emph{arXiv preprint arXiv:2311.16452}, 2023.

\bibitem[{OpenAI}(2022)]{openai2022chatgpt}
{OpenAI}.
\newblock {GPT-3.5}.
\newblock \url{https://openai.com/blog/chatgpt/}, 2022.

\bibitem[OpenAI(2023)]{gpt-4}
OpenAI.
\newblock Gpt-4 technical report.
\newblock \emph{arXiv preprint arXiv:2303.08774}, 2023.

\bibitem[Pal et~al.(2022)Pal, Umapathi, and Sankarasubbu]{pal2022medmcqa}
Pal, A., Umapathi, L.~K., and Sankarasubbu, M.
\newblock Medmcqa: A large-scale multi-subject multi-choice dataset for medical domain question answering.
\newblock In \emph{Conference on Health, Inference, and Learning}, pp.\  248--260. PMLR, 2022.

\bibitem[Pal et~al.(2023)Pal, Umapathi, and Sankarasubbu]{pal2023med}
Pal, A., Umapathi, L.~K., and Sankarasubbu, M.
\newblock Med-halt: Medical domain hallucination test for large language models.
\newblock In \emph{Proceedings of the 27th Conference on Computational Natural Language Learning (CoNLL)}, pp.\  314--334, 2023.

\bibitem[Pudjihartono et~al.(2022)Pudjihartono, Fadason, Kempa-Liehr, and O'Sullivan]{pudjihartono2022review}
Pudjihartono, N., Fadason, T., Kempa-Liehr, A.~W., and O'Sullivan, J.~M.
\newblock A review of feature selection methods for machine learning-based disease risk prediction.
\newblock \emph{Frontiers in Bioinformatics}, 2:\penalty0 927312, 2022.

\bibitem[Qiao et~al.(2024)Qiao, Zhang, Fang, Luo, Zhou, Jiang, Lv, and Chen]{qiao2024autoact}
Qiao, S., Zhang, N., Fang, R., Luo, Y., Zhou, W., Jiang, Y., Lv, C., and Chen, H.
\newblock Autoact: Automatic agent learning from scratch for qa via self-planning.
\newblock In \emph{Proceedings of the 62nd Annual Meeting of the Association for Computational Linguistics (Volume 1: Long Papers)}, pp.\  3003--3021, 2024.

\bibitem[Sackett(1997)]{sackett1997evidence}
Sackett, D.~L.
\newblock Evidence-based medicine.
\newblock \emph{Seminars in perinatology}, 21\penalty0 (1):\penalty0 3--5, 1997.

\bibitem[Sellergren et~al.(2025)Sellergren, Kazemzadeh, Jaroensri, Kiraly, Traverse, Kohlberger, Xu, Jamil, Hughes, Lau, et~al.]{sellergren2025medgemma}
Sellergren, A., Kazemzadeh, S., Jaroensri, T., Kiraly, A., Traverse, M., Kohlberger, T., Xu, S., Jamil, F., Hughes, C., Lau, C., et~al.
\newblock Medgemma technical report.
\newblock \emph{arXiv preprint arXiv:2507.05201}, 2025.

\bibitem[Singhal et~al.(2023)Singhal, Tu, Gottweis, Sayres, Wulczyn, Hou, Clark, Pfohl, Cole-Lewis, Neal, Schaekermann, Wang, Amin, Lachgar, Mansfield, Prakash, Green, Dominowska, y~Arcas, Tomasev, Liu, Wong, Semturs, Mahdavi, Barral, Webster, Corrado, Matias, Azizi, Karthikesalingam, and Natarajan]{medpalm2}
Singhal, K., Tu, T., Gottweis, J., Sayres, R., Wulczyn, E., Hou, L., Clark, K., Pfohl, S., Cole-Lewis, H., Neal, D., Schaekermann, M., Wang, A., Amin, M., Lachgar, S., Mansfield, P., Prakash, S., Green, B., Dominowska, E., y~Arcas, B.~A., Tomasev, N., Liu, Y., Wong, R., Semturs, C., Mahdavi, S.~S., Barral, J., Webster, D., Corrado, G.~S., Matias, Y., Azizi, S., Karthikesalingam, A., and Natarajan, V.
\newblock Towards expert-level medical question answering with large language models.
\newblock \emph{arXiv preprint arXiv:2305.09617}, 2023.

\bibitem[Soroush et~al.(2024)Soroush, Glicksberg, Zimlichman, Barash, Freeman, Charney, Nadkarni, and Klang]{soroush2024large}
Soroush, A., Glicksberg, B.~S., Zimlichman, E., Barash, Y., Freeman, R., Charney, A.~W., Nadkarni, G.~N., and Klang, E.
\newblock Large language models are poor medical coders—benchmarking of medical code querying.
\newblock \emph{NEJM AI}, 1\penalty0 (5):\penalty0 AIdbp2300040, 2024.

\bibitem[Tang et~al.(2024)Tang, Zou, Zhang, Li, Zhao, Zhang, Cohan, and Gerstein]{tang2024medagents}
Tang, X., Zou, A., Zhang, Z., Li, Z., Zhao, Y., Zhang, X., Cohan, A., and Gerstein, M.
\newblock Medagents: Large language models as collaborators for zero-shot medical reasoning.
\newblock In \emph{Findings of the Association for Computational Linguistics: ACL 2024}, pp.\  599--621, 2024.

\bibitem[Toma et~al.(2023)Toma, Lawler, Ba, Krishnan, Rubin, and Wang]{toma2023clinical}
Toma, A., Lawler, P.~R., Ba, J., Krishnan, R.~G., Rubin, B.~B., and Wang, B.
\newblock Clinical camel: An open-source expert-level medical language model with dialogue-based knowledge encoding.
\newblock \emph{arXiv preprint arXiv:2305.12031}, 2023.

\bibitem[Touvron et~al.(2023{\natexlab{a}})Touvron, Lavril, Izacard, Martinet, Lachaux, Lacroix, Rozi{\`e}re, Goyal, Hambro, Azhar, et~al.]{llama}
Touvron, H., Lavril, T., Izacard, G., Martinet, X., Lachaux, M.-A., Lacroix, T., Rozi{\`e}re, B., Goyal, N., Hambro, E., Azhar, F., et~al.
\newblock Llama: Open and efficient foundation language models.
\newblock \emph{arXiv preprint arXiv:2302.13971}, 2023{\natexlab{a}}.

\bibitem[Touvron et~al.(2023{\natexlab{b}})Touvron, Martin, Stone, Albert, Almahairi, Babaei, Bashlykov, Batra, Bhargava, Bhosale, et~al.]{touvron2023llama2}
Touvron, H., Martin, L., Stone, K., Albert, P., Almahairi, A., Babaei, Y., Bashlykov, N., Batra, S., Bhargava, P., Bhosale, S., et~al.
\newblock Llama 2: Open foundation and fine-tuned chat models.
\newblock \emph{arXiv preprint arXiv:2307.09288}, 2023{\natexlab{b}}.

\bibitem[Ullah et~al.(2024)Ullah, Parwani, Baig, and Singh]{ullah2024challenges}
Ullah, E., Parwani, A., Baig, M.~M., and Singh, R.
\newblock Challenges and barriers of using large language models (llm) such as chatgpt for diagnostic medicine with a focus on digital pathology--a recent scoping review.
\newblock \emph{Diagnostic pathology}, 19\penalty0 (1):\penalty0 43, 2024.

\bibitem[Van~Veen et~al.(2024)Van~Veen, Van~Uden, Blankemeier, Delbrouck, Aali, Bluethgen, Pareek, Polacin, Reis, Seehofnerov{\'a}, et~al.]{van2024adapted}
Van~Veen, D., Van~Uden, C., Blankemeier, L., Delbrouck, J.-B., Aali, A., Bluethgen, C., Pareek, A., Polacin, M., Reis, E.~P., Seehofnerov{\'a}, A., et~al.
\newblock Adapted large language models can outperform medical experts in clinical text summarization.
\newblock \emph{Nature Medicine}, pp.\  1--9, 2024.

\bibitem[Wang et~al.(2024)Wang, Ma, Feng, Zhang, Yang, Zhang, Chen, Tang, Chen, Lin, et~al.]{wang2024survey}
Wang, L., Ma, C., Feng, X., Zhang, Z., Yang, H., Zhang, J., Chen, Z., Tang, J., Chen, X., Lin, Y., et~al.
\newblock A survey on large language model based autonomous agents.
\newblock \emph{Frontiers of Computer Science}, 18\penalty0 (6):\penalty0 186345, 2024.

\bibitem[Wu et~al.(2024)Wu, Lin, Zhang, Zhang, Xie, and Wang]{wu2024pmc}
Wu, C., Lin, W., Zhang, X., Zhang, Y., Xie, W., and Wang, Y.
\newblock Pmc-llama: toward building open-source language models for medicine.
\newblock \emph{Journal of the American Medical Informatics Association}, 31\penalty0 (9):\penalty0 1833--1843, 2024.

\bibitem[Yang et~al.(2024)Yang, Xu, Sellergren, Kohlberger, Zhou, Ktena, Kiraly, Ahmed, Hormozdiari, Jaroensri, et~al.]{yang2024advancing}
Yang, L., Xu, S., Sellergren, A., Kohlberger, T., Zhou, Y., Ktena, I., Kiraly, A., Ahmed, F., Hormozdiari, F., Jaroensri, T., et~al.
\newblock Advancing multimodal medical capabilities of gemini.
\newblock \emph{arXiv preprint arXiv:2405.03162}, 2024.

\bibitem[Yao et~al.(2023)Yao, Zhao, Yu, Du, Shafran, Narasimhan, and Cao]{yao2023react}
Yao, S., Zhao, J., Yu, D., Du, N., Shafran, I., Narasimhan, K., and Cao, Y.
\newblock React: Synergizing reasoning and acting in language models.
\newblock In \emph{International Conference on Learning Representations (ICLR)}, 2023.

\bibitem[Zheng et~al.(2024)Zheng, Zhang, Zhang, Ye, Luo, Feng, and Ma]{zheng2024llamafactory}
Zheng, Y., Zhang, R., Zhang, J., Ye, Y., Luo, Z., Feng, Z., and Ma, Y.
\newblock Llamafactory: Unified efficient fine-tuning of 100+ language models.
\newblock In \emph{Proceedings of the 62nd Annual Meeting of the Association for Computational Linguistics (Volume 3: System Demonstrations)}, Bangkok, Thailand, 2024. Association for Computational Linguistics.
\newblock URL \url{http://arxiv.org/abs/2403.13372}.

\bibitem[Zhou et~al.(2023)Zhou, Liu, Gu, Zou, Huang, Wu, Li, Chen, Zhou, Liu, et~al.]{zhou2023survey}
Zhou, H., Liu, F., Gu, B., Zou, X., Huang, J., Wu, J., Li, Y., Chen, S.~S., Zhou, P., Liu, J., et~al.
\newblock A survey of large language models in medicine: Progress, application, and challenge.
\newblock \emph{arXiv preprint arXiv:2311.05112}, 2023.

\end{thebibliography}
\bibliographystyle{icml2026}

\newpage
\appendix
\onecolumn
\section{Framework}
\label{appen:method}

\subsection{Environment}

\paragraph{Environment 1}
identifies the most suitable tool from the tool library for subsequent calculations. 
In our work, we collect 387 tools $T=\{t_1, t_2, ..., t_M\}$, where $M=387$ to form our tool library.
To optimize tool selection, we present an embedding-based retrieval-ranking algorithm, which helps the method accurately and efficiently select the most appropriate tools relevant to patients from the massive available tools, thereby increasing overall model performance.

\textit{Retrieval.}
We first adopt the text-embedding-ada-002 model \cite{text-embedding-ada-002} to extract the embeddings $E_T$ of all tools $T$ using their tool metadata including both tool names and descriptions, with the dimension of 1,536, defined as follows:
\begin{equation}
    E_T = \text{Embedding}(T)
\end{equation}

Then, we adopt the concatenation of patient information and risk questions to represent each patient. We again use the same embedding model to extract the embeddings $E_P$ for all patients $P$. 
\begin{equation}
    E_P = \text{Embedding}(P)
\end{equation}

Finally, we adopt the widely-used cosine similarity to calculate the similarity scores $S$ between patients and all available tools.
\begin{equation}
\begin{aligned}
    S =& \text{Similarity}(T, P) \\ 
    =& \text{Cosine\_Similarity}(E_T, E_P) \\
    =& \frac{E_T \cdot E_P}{\|E_T\| \|E_P\|}
\end{aligned}
\end{equation}

where $(\cdot)$ represents the dot product of the two vectors; and
\( \|\cdot\| \) is the Euclidean norms (i.e., \( L_2 \) norms) of the vector.
Therefore, we can set a threshold and retrieve all relevant tools with a similarity score greater than it.

\textit{Rank.}
In our work, we propose ranking the tools using the calculated similarity scores $S$ and forwarding the top-$N$ tools with the highest similarity scores for the Decider to perform tool selection.
We evaluate the performance of setting different values of $N$.
Our preliminary results show that setting $N=5$ achieves a 99.8\% recall score, meaning that in 99.8\% of cases, the correct tool is included among the top-5 tools retrieved and ranked by our method.
This proves the effectiveness of the embedding-based retrieval-ranking algorithm used.

\paragraph*{Environment 2}
provides structured parameter data so that an external medical tool can be invoked accurately. Once the tool is selected in Environment 1, this environment defines the required schema—including parameter names, types (numeric or categorical), and allowable value options—and ensures that each parameter in the patient information is extracted and validated in the correct format. Specifically, numeric fields must be cast appropriately, and categorical or boolean fields must match the tool’s defined options. This process guarantees that all required parameters are provided in a proper format, thereby minimizing the risk of miscalculation or incompatibility when the tool is invoked. To address the challenge of parsing inconsistencies in raw patient data, such as different unit systems, we provide a comprehensive input schema within our prompts. This schema explicitly specifies the required unit format (e.g., kilograms vs. pounds) and acceptable value ranges for each parameter. Additionally, we implemented a robust retry mechanism to handle potential parameter extraction or formatting errors. When inconsistencies are detected, the system prompts the LLM to re-analyze the patient data, guiding it to correctly extract and format the values according to the required specifications.

\begin{figure*}[!ht]
    \centering
    \begin{tikzpicture}
    \node[draw=black, 
    line width=1pt,
    rounded corners=10pt,
    text width=\textwidth,
    inner sep=5pt,
    align=left](box){
        \small
        \textbf{a) Environment 1}
         
        Select the most appropriate assessment tool for the following case and question:{\{Patient Information and Question\}}\\
        Available Tools:\{List of Available Tools and Descriptions\}\\        
        Please respond only one tool that fits the question best in this format:\\        
        Tool\_xx. [tool name]\\
        Analysis: Brief justification in 2-3 sentences.

        \noindent\rule{\dimexpr1\textwidth}{0.4pt}
        
        \textbf{b) Environment 2}
        
        Analyze the medical case and output parameters based on the schema.

        Patient case:
        {\{Patient Information\}}\\
        RULES:\\
        1. Output format MUST be: \{"name": value\}\\
        2. Use EXACT "name" fields from schema as keys\\
        3. Include ALL fields from schema\\
        4. Use ONLY values defined in schema options\\
        Schema:
        \{Input Schema\}
        
        \noindent\rule{\dimexpr1\textwidth}{0.4pt}
        
        \textbf{c) Environment 3}
        
        Based on the calculator's output:
        \{Calculator Result JSON\}\\
        Please conclude the results and answer the question:
        {\{Question\}}
        
        \noindent\rule{\dimexpr1\textwidth}{0.4pt}
        \textbf{d) Environment 4}
        
        Based on the analysis:
        \{Interpretation Results\}

        Select the best answer for the question {\{Question\}}:\\
        A) \{Option A\}\\
        B) \{Option B\}\\
        C) \{Option C\}\\
        D) \{Option D\}\\

        Respond with format: Finish[A/B/C/D]

        \noindent\rule{\dimexpr1\textwidth}{0.4pt}
        
        \textbf{e) Environment 5}
        
        Validate the following clinical case analysis stages for correctness:\\
        Case Question: \{Question with Patient Information\}

        1. Tool Selection:
        Input: \{ENV1 Input\}; Output: \{ENV1 Output\}

        2. Parameter Extraction:
        Input: \{ENV2 Input\}; Output: \{ENV2 Output\}

        3. Result Interpretation:
        Input: \{ENV3 Input\}; Output: \{ENV3 Output\}

        4. Answer Selection:
        Input: \{ENV4 Input\}; Output: \{ENV4 Output\}

        Required Output Format:\\
        RESULT: Reflect[RIGHT/WRONG]\\
        ANALYSIS:
        "All stages processed correctly" / Stage\_X: [ERROR] <error description>

        Instructions:\\
        1. Response start with Reflect[RIGHT] or Reflect[WRONG] with ANALYSIS strictly follow the required format\\
        2. If error found, report only the earliest error stage with justification
        
    };
    \end{tikzpicture}
    \caption{The instructions and prompts used in our three introduced LLM-based agents.} 
    \label{fig:prompts} 
\end{figure*}

\paragraph{Environment 3}
manages how tool-generated results are interpreted and rephrased in light of the original patient information. After valid parameters are provided from Environment 2, the system invokes the chosen medical tool (e.g., DECAF Score) to calculate a risk level and yield a concise statement (e.g., ``31\% in-hospital mortality''). Environment 3 then transforms this raw output into a coherent free-text explanation that references both the patient information and the clinical question. For instance, a high-risk DECAF Score might be rephrased as: ``A DECAF score of 4 places patients in a high-risk category, correlating with a mortality rate near 31\%.'' The Decider can then utilize this refined statement to guide higher-level decisions or generate a final answer.

\paragraph{Environment 4}
focuses on transforming high-level textual conclusions into standardized outputs, such as multiple-choice answers or discretized risk scores. When the Decider generates a preliminary conclusion in free-text form, this environment formats it into a concise, unambiguous response. For instance, in a multiple-choice setting, it explicitly requests the system to produce an answer in the form \texttt{Finish[A/B/C/D]}. This requirement helps ensure that the final output is straightforward to parse and evaluate, making it suitable for automated scoring against ground-truth labels.

\paragraph{Environment 5}
In the final stage of the pipeline, Environment 5 collects all historical analysis and system outputs, enabling the Reviewer to perform a global verification. The Reviewer checks the correctness of each step. If an error is detected, the Reviewer pinpoints the earliest failing environment and provides revision instructions. The system then re-executes from that stage, incorporating the feedback. This iterative cycle continues up to three attempts; if errors still persist beyond this limit, the process terminates with the last available result. Such a mechanism maintains overall computational efficiency while still allowing multiple opportunities to correct reasoning flaws.

\subsection{LLM Agents}
\paragraph{Decider}

The Decider is tasked with making key clinical decisions throughout the workflow. It first utilizes Environment 1 to identify and select the most contextually relevant tool, and then interacts with Environment 3 to interpret the outputs from that tool. 
Figure~\ref{fig:prompts} (a) and (c) show the detailed prompts used in Reviewer. 
Leveraging the patient’s information and the tool’s computed results, the Decider produces a synthesized answer addressing the posed clinical question. This structured delegation—where the Decider focuses on reasoning and choice of tools—enhances clarity and reduces the complexity of subsequent tasks.

\paragraph{Executor}

The Executor’s role is to carry out the instructions determined by the Decider precisely. After a specific tool is selected, the Executor references Environment 2 to extract and validate each input parameter. 
It then calls the tool externally and obtains its results. Figure~\ref{fig:prompts} (b) illustrates the used system prompts.
Finally, as shown in Figure~\ref{fig:prompts} (d), the Executor consults Environment 4 to format the system’s textual output into a predetermined style (e.g., a discrete choice or numerical score).

\paragraph{Reviewer}
The Reviewer, leveraging information stored in Environment 5, ensures the correctness and reliability of the entire pipeline. Once the Executor produces a final answer, the Reviewer meticulously inspects each environment’s input and output history. If any discrepancy or error is identified (e.g., using the wrong tool, misformatted parameters, or an incorrect conclusion), the Reviewer indicates the earliest environment that requires revision and provides targeted feedback. The system then restarts from that stage, integrating the Reviewer’s suggestions. 
Figure~\ref{fig:prompts} (e) details the prompts used in Reviewer.

\section{Settings and Baselines}

\subsection{Settings}

We perform model fine-tuning using LoRA\cite{hu2022lora} following the approach introduced using the LLaMA Factory framework \cite{zheng2024llamafactory}. The MedRisk benchmark dataset is divided in a 7:1:2 ratio for training, validation, and testing, respectively. All training is conducted using PyTorch's \textit{DistributedDataParallel} on 4$\times$A100 (80GB) GPUs with BF16 precision. We use AdamW optimizer \cite{loshchilov2019adamW} with a cosine learning rate scheduler. The effective batch size is 32 (batch size 8 per device with 4 gradient accumulation steps). The detailed hyper-parameters for training are shown in Table~\ref{tab:hyper_parameters}.

\begin{table}[h]
\centering
\footnotesize
\caption{
We illustrate the detailed hyper-parameters for model training.
}
\label{tab:hyper_parameters}
\setlength{\tabcolsep}{3pt} 
\begin{tabular}{@{}lc@{}}
\toprule
\textbf{Hyper\_parameters}   &  \textbf{Value}
\\ 
\midrule

GPUs & 4*A100 (80GB)
\\ 

epochs & 5
\\

learning\_rate & 0.0001
\\

model\_max\_length & 2,560
\\

per\_device\_batch\_size & 8
\\ 

gradient\_accumulation\_steps & 4
\\

batch size & 8*4*4
\\

warmup\_ratio & 0.1
\\

optimizer &  AdamW 
\\

lr\_scheduler & cosine
\\

precision & bf16 
\\

finetuning\_method & LoRA 
\\

lora\_r    &  8  
\\

lora\_alpha & 16 
\\

lora\_dropout & 0
\\

lora\_target\_modules & all

\\
\bottomrule
\end{tabular}
\end{table}

During evaluation, for a fair comparison, we test all models on the same test set using a consistent protocol. We use a temperature of 0.6 and top-p of 0.9 for controlled randomness in generation. Each model receives identical input formatting, consisting of patient information, a clinical question, and four multiple-choice options. The metrics focus on accuracy, calculated as the exact match between model predictions and ground truth answers.

\subsection{Baselines}
\label{appen:baselines}
To comprehensively evaluate LLMs' performance for risk prediction, we collect 19 different types of representative LLMs and agents.
1) \textit{Public LLMs}: We collect multiple representative open-source public LLMs across different numbers of model parameters, i.e., Mistral-7B \cite{jiang2023mistral}, LLaMA-3-8B \cite{llama3modelcard}, Mixtral-8x7B \cite{jiang2024mixtral}, and LLaMA-3-70B \cite{llama3modelcard}; We also collect three public medical LLMs:
PMC-LLaMA-13B \cite{wu2024pmc}, Meditron-70B  \cite{chen2023meditron}, and MedGemma \cite{sellergren2025medgemma}. They are developed by fine-tuning general LLMs on a large corpus of medical datasets \cite{zhou2023survey} to learn rich medical knowledge, achieving improved performance in medical tasks.
2) \textit{Commercial LLMs}: We further collect the currently most powerful commercial LLMs, 
GPT-3.5 \cite{openai2022chatgpt}, GPT-4 \cite{gpt-4},
GPT-4o-2024-08-06 \cite{hurst2024gpt4o}, o3-mini-2025-01-31 \cite{gpt2025o3}, o1-2024-12-17 \cite{jaech2024openaio1}, GPT-4.5 \cite{gpt_4.5}, Claude-3.7-Sonnet \cite{Claude37Sonnet}, Claude-4-Sonnet \cite{Claude4Sonnet}, and Claude-4-Opus \cite{Claude4Opus}.
3) \textit{AI Agents}: Recently, AI agents \cite{wang2024survey,qiao2024autoact}, which assign different LLMs to different roles to collaboratively perform tasks, have demonstrated better task performance than single LLMs in traditional domains such as gaming and programming. Therefore, for a comprehensive comparison, we further collect three well-known and recent AI agents, i.e., ReAct \cite{yao2023react}, BOLAA \cite{liu2023bolaa}, and Chameleon \cite{lu2024chameleon}.
We adopt LLaMA-3-8B \cite{llama3modelcard} as the backbone of our system to re-implement them for a fair comparison.

\begin{figure*}[t]
\centering
\includegraphics[width=1\linewidth]{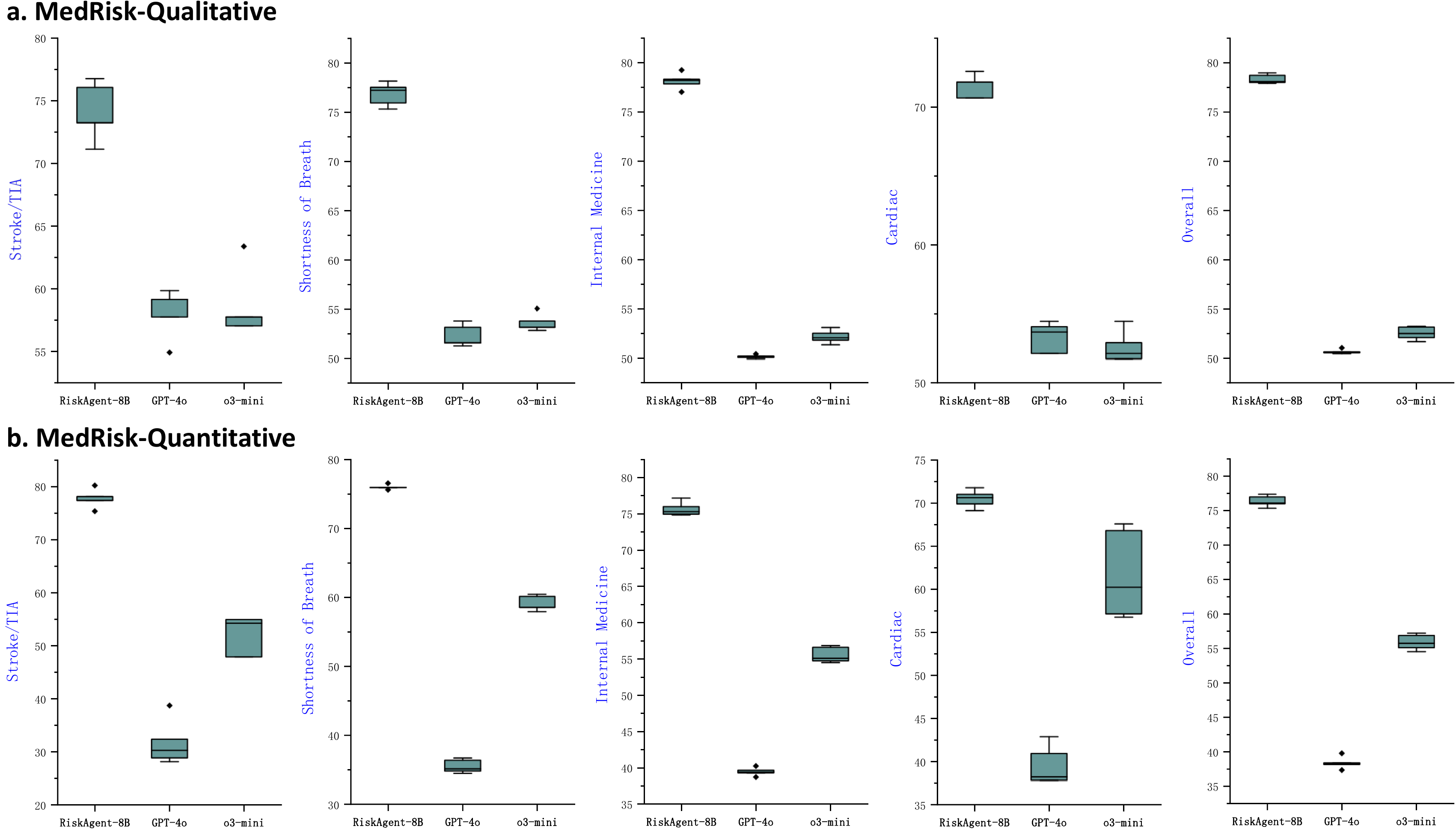}
\caption{Performance for RiskAgent-8B, GPT-4o, and o3-mini. 
In the boxplot, the central line indicates the median value, while the lower and upper boundaries represent the 25th (Q1) and 75th (Q3) percentiles, respectively. 
The whiskers extend up to 1.5 times the interquartile range (IQR).
We can see that RiskAgent-8B not only consistently achieves the best results across all cases in different runs but also achieves lower perturbations (STD) than o3-mini in most cases, especially in risk prediction for cardiac.}
\label{fig:boxplot}
\end{figure*}

\end{document}